\documentclass{ifacconf}

\usepackage{graphicx}      
\usepackage{natbib}  
\usepackage{amsmath,amssymb,amsfonts}
\usepackage{float}
\usepackage{algorithm}
\usepackage{algcompatible}
\usepackage{algpseudocode}
\usepackage{graphicx}
\usepackage{textcomp}
\usepackage{xcolor}
\usepackage{xcolor}
\usepackage{optidef}
\usepackage{subfiles}
\usepackage{caption}

\usepackage{subcaption}
\usepackage[export]{adjustbox}

\def\BibTeX{{\rm B\kern-.05em{\sc i\kern-.025em b}\kern-.08em
    T\kern-.1667em\lower.7ex\hbox{E}\kern-.125emX}}

\newcommand{\eg}{\textit{e.g.,}}

\begin{document}
\begin{frontmatter}

\title{Distributed Optimal Control Framework for High-Speed Convoys: Theory and Hardware Results} 


\author[First]{Namya Bagree}
\author[Second]{Charles Noren}
\author[First]{Damanpreet Singh}
\author[Second]{Matthew Travers}
\author[Second]{Bhaskar Vundurthy}

\address[First]{Department of Mechanical Engineering, Carnegie Mellon University, USA (e-mail: \{nbagree, damanprs\}@andrew.cmu.edu).}
\address[Second]{The Robotics Institute, Carnegie Mellon University, USA (e-mail: \{cnoren, mtravers, pvundurt\}@andrew.cmu.edu).}

\begin{abstract}                
Practical deployments of coordinated fleets of mobile robots in different environments have revealed the benefits of maintaining small distances between robots, especially as they move at higher speeds. 
However, this is counter-intuitive in that as speed increases, reducing the amount of space between robots also reduces the time available to the robots to respond to sudden motion variations in surrounding robots. 
However, in certain examples, the benefits in performance due to traveling at closer distances can outweigh the potential instability issues, for instance, autonomous trucks on highways that optimize energy by vehicle ``drafting'' or smaller robots in cluttered environments that need to maintain close, line of sight communication, etc. 
To achieve this kind of closely coordinated fleet behavior, this work introduces a model predictive optimal control framework that directly takes non-linear dynamics of the vehicles in the fleet into account while planning motions for each robot. 
The robots are able to follow each other closely at high speeds by proactively making predictions and reactively biasing their responses based on state information from the adjacent robots.
This control framework is naturally decentralized and, as such, is able to apply to an arbitrary number of robots without any additional computational burden. 
We show that our approach is able to achieve lower inter-robot distances at higher speeds compared to existing controllers.
We demonstrate the success of our approach through simulated and hardware results on mobile ground robots. 
\end{abstract}


\end{frontmatter}

\section{Introduction} \label{sec: introduction}
Fleets of mobile robots have shown practical benefits by moving closer to each other at high speeds to optimize system resource utilization and improve process efficiencies. An example of this is a group of trucks that reduce fuel consumption through the reduction of aerodynamic drag by following at lower inter-vehicle distances (\citealp{Nahavandi}; \citealp{Turri}). However, reducing inter-vehicle distances reduces the amount of time available to respond to sudden unexpected behaviors or motions of the other robots in the fleet. This can lead to instabilities in the behavior of the overall system. To address these problems, our work investigates non-linear control techniques for autonomous robots that can maintain low inter-robot distance over various environments at high speeds. 

While the advantages of a fleet of robots exist in many domains (\citealp{Nahavandi}), small mobile robots can also see performance improvements by driving fast at close distances in unstructured regions. A practical challenge in search and rescue missions is maintaining consistent communication between the robots, especially in cluttered environments where the loss of direct line of sight can hinder inter-robot communications (\citealp{travers2022}). This is another example of a scenario where actively maintaining low inter-robot separation distances is essential to the performance of the team. To this end, in this work, we deploy such a small-scale system in both simulations as well as hardware.

\begin{figure}[!t]
\centering
\includegraphics[width=3.49in]{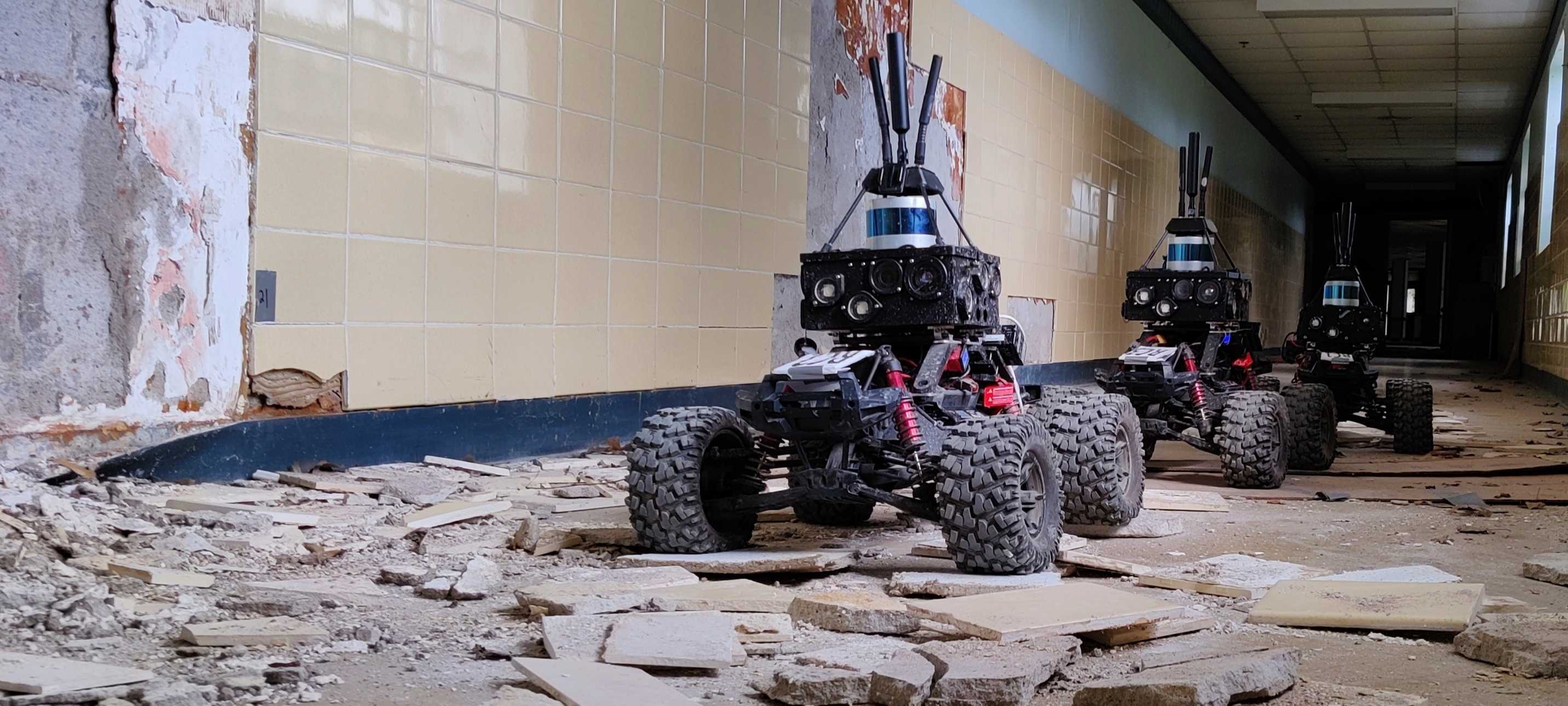}
\caption{Robot convoy performing a search and rescue task}
\label{fig:conv_rubble}
\end{figure}

The primary focus of this work is to present a non-linear model based controller that addresses many issues with the current state-of-the-art methods for controlling fleets of mobile robots operating in cluttered environments. Our controller takes information from all of the adjacent robots and computes both the feedforward as well as the feedback components that help it proactively plan as well as reactively adjust behavior to compensate for any unexpected behaviors across the fleet. 

In addition, we designed the control framework to operate in a decentralized fashion. This allows us to extend our framework to an arbitrary number of robots with a constant computational requirement for each robot. We demonstrate the overall efficacy of our approach via experiments in simulation as well as hardware, for a fleet of wheeled ground vehicles.

\section{Problem Definition} \label{sec:probstatement}

\begin{figure}[t!]
\centering
\includegraphics[width=0.45\textwidth]{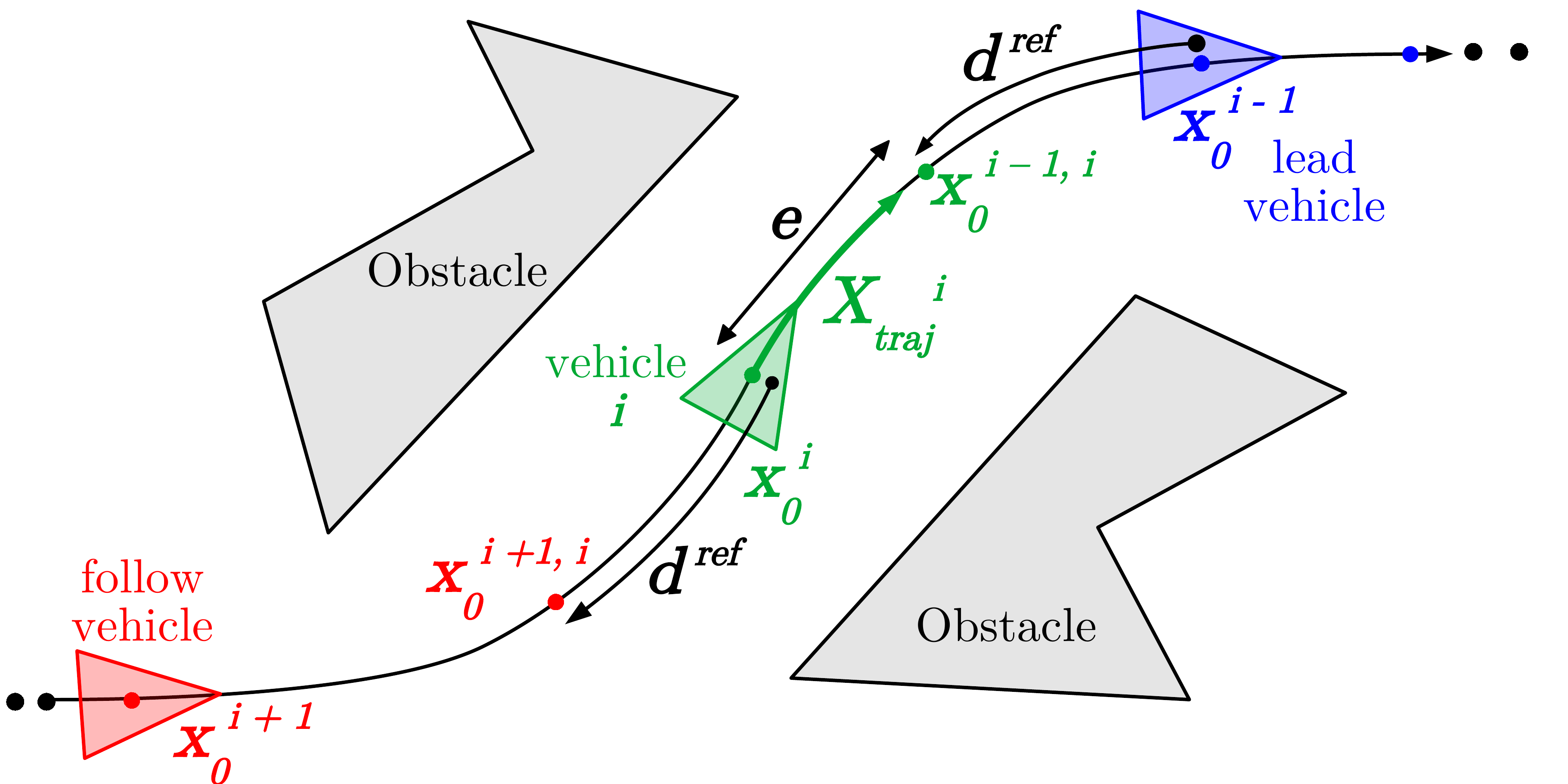}
\caption{Schematic for the convoy control problem. Lead and follow vehicles are defined with respect to vehicle $i$. Error $e$ is the Euclidean distance between a vehicle's desired position ($d^{ref}$ behind the lead vehicle) and its current location. }
\label{fig:problem_schematic}
\end{figure}

Consider the schematic illustrated in Fig. \ref{fig:problem_schematic} for vehicle $i$. We assume an increasing index along the trajectory and thus refer to the neighboring vehicles as the lead vehicle and the follow vehicle denoted by $i-1$ and $i+1$, respectively. Given $L$ agents in the fleet, the state and control trajectories for the agent $i$ over a time horizon of length $N$ are given by $X^i = [x_0^i, x_1^i, ..., x_N^i]$ and $U^i = [u_0^i, u_1^i, ..., u_{N-1}^i]$, respectively, where $i \in \mathcal{I}_L = \{1, 2, \cdots, L\}$. 

The objective of the convoy problem as we define it in this work is to design a distributed convoy controller that generates a control output $u^i_0$ at every time instant to get a fleet of robots to follow a predefined trajectory. In particular, the desired objectives for each vehicle are to

\begin{enumerate}
    \item Maintain a fixed distance $d^{ref}$ along the desired path from its lead vehicle; 
    \item Simultaneously travel at desired speeds;
    \item React to disturbances without creating instabilities to the overall behavior of the convoy.
\end{enumerate}

\section{Literature Review} \label{LitReview}

To provide context for our control approach, we review relevant prior works that address the convoy control problem defined in Section \ref{sec:probstatement}. Despite several use cases for convoys (\citealp{Nahavandi}), we focus our discussion on works for unstructured and cluttered environments.

One of the earliest works (\citealp{Yazbeck}) in convoy control mimics a leader-follower behavior where each vehicle estimates and stores the path of its predecessor as a set of points. The follower then estimates the predecessor's path curvature around a selected target and follows the trajectory. (\citealp{Nestlinger}) extend this work to store position measurements over time and apply a spline-approximation technique to obtain a smooth reference path for the underlying motion controllers. However, these methods force the robot to track the exact positions of its predecessors, restricting system flexibility around obstacles.

\citealp{Albrecht} provides a framework to switch between exact pose tracking and flexible path search and tracking based on the environment. This allows the trajectory following convoy algorithms to operate in real-world conditions. However, there is no interaction between the outputs of the two tracking methods, reducing the controller performance due to conflicting switch decisions when operating in a cluttered environment.

In contrast to (\citealp{Yazbeck}; \citealp{Nestlinger}; \citealp{Albrecht}), there exist prior works of convoy control wherein they specifically provide an integrated obstacle avoidance module to operate a fleet of robots in unstructured environments. For instance, in (\citealp{Zhao}), the authors define a robust solution using an adaptive inter-robot distance control and leader pose ($x_0^{i-1}$) estimation. However, there are no common velocity-based control parameters and no prediction on future paths for adjacent agents ($X^{i-1}, X^{i+1}$). This leads to higher angular variation as inter-robot distances reduce, resulting in wavy motions and higher error in tracking at high speeds.

An additional prior work that incorporates specific obstacle avoidance in the control framework is presented by (\citealp{Shin}). Here, the authors approach the convoy problem with a controller that incorporates a passivity-based MPC method that explicitly integrates a traversability map in the planner. However, this framework doesn't take feedback from the following vehicles ($i+1$), which can lead to high inter-robot distances on cluttered terrains. They also rely on continuous communication between the vehicles and a base node to operate. 

One other work that doesn't apply the ``follow-the-leader'' framework as discussed earlier, is presented by (\citealp{Turri}). The authors in this work take a centralized approach to designing their controller, focusing primarily on straight-line velocity profiles. The objective of the optimization in their work is to improve fuel efficiency while ensuring desired safe following distances between vehicles. The computational cost increases quadratically as you add new agents and the overall approach needs to be solved on a single computer, creating a single point of failure. This also requires high-rate continuous communications between the robots and the base node for safe convoying. 

\section{Convoy Behavior} \label{sec:ConvoyBehvaior}

\begin{figure}[t!]
     \centering
     \begin{subfigure}[b]{0.45\textwidth}
         \centering
         \includegraphics[width=\textwidth]{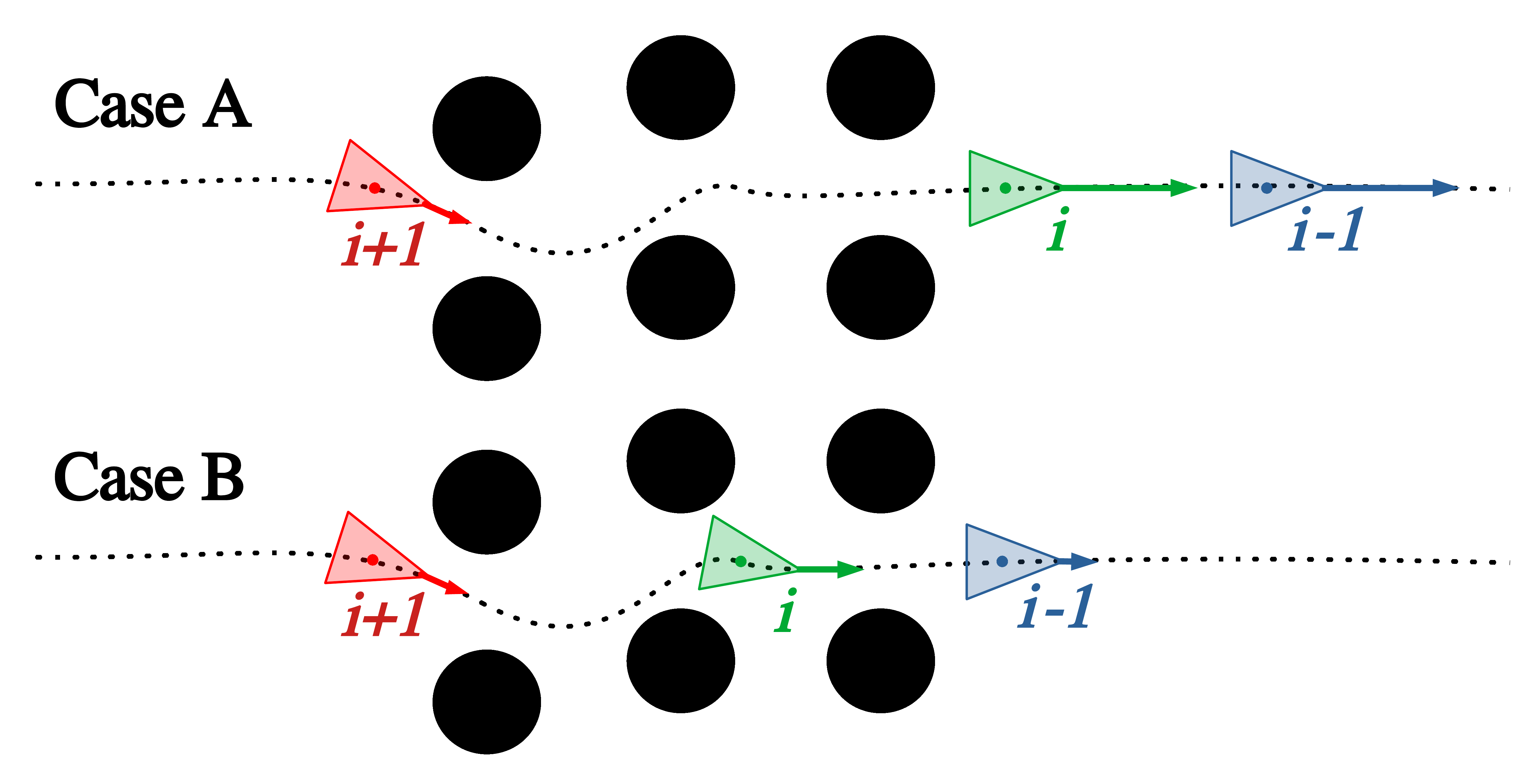}
         \caption{Environmental situation where robot $i+1$ is stuck}
         \label{fig:lit_review_env}
     \end{subfigure}
     \hfill
     \begin{subfigure}[b]{0.3\textwidth}
         \centering
         \includegraphics[width=\textwidth]{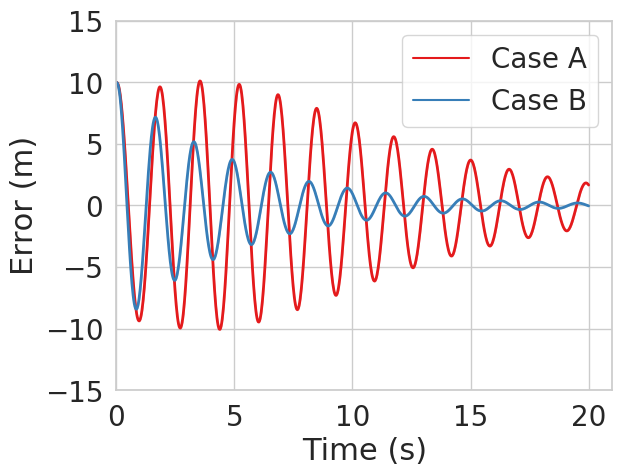}
         \caption{Fleet response demonstrating the accordion problem}
         \label{fig:spring_damper}
     \end{subfigure}
        \caption{Comparison of convoys when following robots are neglected in the framework (Case A) vs when they are not (Case B).}
        \label{fig:lit_review}
\end{figure}

Our convoy system is designed to include the best aspects of an explicit ``follow-the-leader'' behavior and a completely centralized controller. We display a scenario in Fig.~\ref{fig:lit_review_env} to better understand the comparison of a ``follow-the-leader'' (Case A) behavior against ours (Case B). Agent $i+1$ has been forced to slow down while navigating a cluttered environment. A spring-damper-mass system was defined for three agents to simulate both behaviors. Each agent is only affected by its lead agent in Case A and both its adjacent agents in Case B. Fig.~\ref{fig:spring_damper} shows that Case B stabilizes faster and reduces accordion-like effects between the agents. This also ensures lower inter-robot distances through the environment, assisting real-time communication between the agents. 

The high-level controller behaviors listed in Section \ref{sec:probstatement} can be expressed in an optimal control framework. The optimal control framework presented in this work expresses the behaviors described in~(1) and~(2) as terms in the controller objective. The proposed controller's cost structure encodes agent reactivity to disturbances described in behavior~(3). We now describe the proposed controller in detail.

\subsection{Convoy Controller} \label{subsec:c_controller}
For the $i$th robot in the convoy, the discrete-time optimal control problem framework is posed as:
\begin{mini}
	{\forall i \in \mathcal{I}_L} {C_{traj}(X^i, U^i) + C_{convoy}(X^{i}, X^{i-1}, X^{i+1})}{}{}
	\addConstraint{x^i_{k+1} = f(x^i_k,u^i_k, \Delta t), \; }{\forall k = 0,...,N-1} 
	\addConstraint{x^i_0=x^i(0), u^i_0=u^i(0), \; }{\forall i \in {I}_L},
\end{mini} 
where:
\begin{equation} \label{eqn:a1_c_cost}
\begin{split}
    &C_{traj}  (X^i, U^i) = \sum_{k=0}^{N-1} (x_k^i-x^i_{traj,k})^TQ(x_k^i-x^i_{traj, k}) \\
             & + u_k^{(i)T}Ru_k^i + (x_N^i-x^i_{traj, N})^TQ_f(x_N^i-x^i_{traj, N}),
\end{split}
\end{equation}
\begin{equation} \label{eqn:c_convoy}
\begin{split}
    C_{convoy} & (X^{i-1}, X^{i}, X^{i+1}) = \\
    & \sum_{k=0}^{N-1} (x^i_k - x^{ref, i-1, i}_k)^T Q_{lead}
    (x^i_k - x^{ref, i-1, i}_k)\\
    & +(x^i_k - x^{ref, i, i+1}_k)^T Q_{follow}
    (x^i_k - x^{ref, i, i+1}_k),
\end{split}
\end{equation}
and where: $x_k \in \mathbb{R}^m$, $u_k\in \mathbb{R}^n$. Additionally, $Q, Q_f, Q_{lead},$ $Q_{follow} \in \mathbb{R}^{m \times m}$ are symmetric positive definite matrices, and $R \in \mathbb{R}^{n \times n}$ is a positive definite matrix. The state-transition model for the $i$th robot is defined as $f(x_k^i, u_k^i, \Delta t)$. Superscript $ref$ refers to the reference point defined between the agents with the indices following $ref$ (\textit{i.e.} either $i-1, i$ or $i, i+1$). The computation of this point is described in the following section. As the run-time cost is evaluated over the same time interval, the run-time costs in \eqref{eqn:a1_c_cost} and \eqref{eqn:c_convoy} may be collapsed into a single quadratic cost expression. This new expression is defined as
\begin{equation} \label{sec4_eqn: full_runtime_cost}
\begin{aligned}
    g(x_k,u_k) =  \text{ }&(x_k^i-Q_T^{-1}y_T)Q_T(x_k^i-Q_T^{-1}y_T) \\
    &\text{ }- y_T^TQ_Ty_T + Z_T+u_k^{(i)T}Ru_k^i,
\end{aligned}
\end{equation}
where:
\begin{align*}
    Q_T &= Q + Q_{lead} + Q_{follow} \\
    y_T &= Qx_{traj, k}^i + Q_{lead}x_{k}^{ref, i-1, i} + Q_{follow}x_k^{ref, i, i+1} \\
    Z_T &= (x_{traj, k}^i)^TQx_{traj, k}^i + (x_{k}^{ref, i-1, i})^TQ_{lead}(x_{k}^{ref, i-1, i}) \\ &+ (x_k^{ref, i, i+1})^TQ_{follow}(x_k^{ref, i, i+1}). 
\end{align*}
This operation is detailed further in Appendix A.
A similar combination of quadratic expressions can be performed on the terminal cost, yielding $(Q_F, y_F, Z_F)$, respectively. These parameters may be used to rephrase the terminal cost:
\begin{equation*}
    \phi(X_N) = (x_N^i-Q_F^{-1}y_F)Q_F(x_N^i-Q_F^{-1}y_F) - y_F^TQ_Fy_F + Z_F.
\end{equation*}
Thus the cost function may be rephrased as:
\begin{equation*}
    J = C_{traj} + C_{convoy} =  \sum_{k=0}^{N-1} \{ g(x_k,u_k) \} + \phi(X_N). 
\end{equation*}
Furthermore, by linearizing the system around $x_k$, \text{ }$u_k$ and defining  $A_k = \frac{\partial}{\partial x_k}f(x_k,u_k)$ and $B_k = \frac{\partial}{\partial u_k}$, the optimization problem can be interpreted and solved online as an iterative Linear Quadratic Regulator~\citealp{iLQR_Todorov}. This yields a control law:
\begin{align*}
    u_k = &-[R_k + B_k^TP_{k+1}B_k]^{-1}B_k^TP_{k+1}A_k(x_k^i-Q_T^{-1}y_T) \\
    =&-K_k(x_k^i-Q_T^{-1}y_T),
\end{align*}
with $P_k$ representing the solution to the Riccati Equation and $K_k$ being the optimal control gain matrix.

\subsection{Controller Implementation and Design Discussion}
The first component of the cost function is a quadratic trajectory tracking cost penalizing deviations from a given convoy trajectory: $x_{traj}$. This cost is defined in \eqref{eqn:a1_c_cost}. In this work, the reference path $x_{traj}$ is provided to the controller as either a pre-defined path or created for each individual robot from observing the motion of other agents in the convoy \eg~``follow-the-leader'' style approaches (\citealp{Nestlinger,Yazbeck}).

The convoy cost $C_{convoy}$ penalizes deviance from the convoy structure over the future time horizon. This cost is defined in~\eqref{eqn:c_convoy} where $x^{ref, i-1, i}_k, x^{ref, i, i+1}_k$ are the reference positions of the $i^{th}$ given the positions of the $i-1$ and $i+1$ cars, respectively, and $Q_{lead}, \text{ } Q_{follow}$ are tunable positive semi-definite constant matrices. The state vector includes the $x$ and $y$ coordinates, vehicle orientation $\psi$ and velocity $v$. The control vector includes acceleration $a$ and steering angle $\delta$.

The computation of $x^{ref}$ is based on the desired inter-robot distance, $d^{ref}$. This desired inter-robot distance is defined as
\begin{equation*}
    d^{ref} = \lambda_1 v_t + \lambda_2 (v_i - v_{i-1}) + K,
\end{equation*}
where $v_t$ is the desired target velocity and $v_i$ corresponds to current velocity for agent $i$. $\lambda_1$ and $\lambda_2$ are tunable parameters where $\lambda_1$ and $\lambda_2$ are non-negative values. $K$ is a constant minimum inter-robot distance for safe operation. Only the current velocities of vehicle $i$ and the prior robot in the convoy structure are considered for this work.

As shown in Fig. \ref{fig:problem_schematic}, the reference positions for agents $i-1$ and $i+1$ over horizon $1:N$ are recovered by performing an open-loop forward rollout using the linearized dynamics at the $i-1$ or $i+1$ agent's state. The $i-1$ and $i+1$ agents' current velocity and steering are assumed to be constant over the rollout. The reference positions for \eqref{eqn:c_convoy} are set by moving backward $d^{ref}$ along the convoy trajectory from the predicted $i-1$ agent positions $X^{i-1}$ and moving the same distance ahead of the $i+1$ agent positions $X^{i+1}$.

The reactivity described in (3) of \ref{sec:probstatement} aims to prevent collisions between agents due to sudden variations in speed. To enable this behavior, the controller computes a weighting factor, $w_{convoy}$, between the costs \eqref{eqn:a1_c_cost} and \eqref{eqn:c_convoy} during run-time. This factor is based on the desired convoy spacing and the current Euclidean distance $dist_{i, j}$ between agents $i$ and $j$. These weights are multiplied to the $Q$ matrices in \eqref{eqn:c_convoy} and the weighting factor is computed as:
\begin{equation*}
    Q_{lead} = w_{i, i-1} * Q_{lead}    
\end{equation*}
\begin{equation*}
    Q_{follow} = w_{i, i+1} * Q_{follow}    
\end{equation*}
\begin{equation*}
    w_{i, j}= 
\begin{cases}
    1 + \frac{w_{far} \times (dist_{i, j} - d^{ref})}{dist_{i, j}} & \text{if } dist_{i, j}\geq d^{ref}\\\
    1 + \frac{w_{near} \times (d^{ref} - dist_{i, j})}{dist_{i, j}} & \text{otherwise.}
\end{cases}
\end{equation*}
The proposed formulation allows the robot to track its predecessor and follower through the predicted $x^{ref}$ terms while also tracking its desired planned path. The proposed additional convoy cost provided in the optimal control formulation may be interpreted as a modification of the local linearization point used in the LQR. This modification of the coordinate transfer from the reference path ($x_{traj}$) through both the user-defined weightings  $(Q, Q_{lead}, Q_{follow})$ and reference trajectories of the leading ($X^{lead}$) and following ($X^{follow}$) robots in the fleet.

\begin{algorithm}
\caption{Convoy controller with obstacle avoidance for robot $i$}\label{alg:convoy_behavior}
\textbf{Input:} Robot states $x_0^{i-1}$, $x_0^i$, $x_0^{i+1}$\\
\textbf{Output:} Control sequence $U^i$
\begin{algorithmic}
\While{\text{Robots are in convoy, $i \in \mathcal{I}_L$}}
\State Run convoy controller, Section \ref{subsec:c_controller}
\If{Output path $X^i$ is obstacle-free}
    \State \Return{Control sequence $U^i$}
\Else
    \State Define $D_{LookAhead}$ based on velocity $v_t$
    \State Calculate desired velocity $v_T$ and direction $\theta_T$
    \State Run the local planner \citealp{Zhang}
    \State Get the modified obstacle-free path $X^i$
    \State Send $X^i$ into a path following iLQR controller
    \State \Return{Control sequence $U^i$}
\EndIf
\EndWhile
\end{algorithmic}
\end{algorithm}

\subsection{Local Planner and Trajectory Controller}

In the absence of an obstacle free path, Algorithm \ref{alg:convoy_behavior} ensures high-speed following and maintains the convoy formation with the help of an additional velocity scaling term:

\begin{equation*}
v_T = (1+\alpha) v_{tc}
\end{equation*}
\begin{equation*}
\alpha =  \lambda_3 (d_1 - d^{ref}) - \lambda_4 d_2
\end{equation*}
Where $\lambda$'s are tunable parameters, $v_{tc}$ is the desired target velocity from the convoy controller, $v_T$ is the modified desired target velocity for the planner, $v_r$ is the robot's velocity, $v_l$ is the leader's velocity, $d_1$ is the distance between the robot and the leader along the trajectory and $d_2$ is the distance between the robot and the follower along the trajectory.

The desired direction is selected based on the output of the convoy controller. This is done to track the desired controller path to the greatest extent before the robot switches back into the convoy controller mode. The direction $\theta_T$ is defined via the look ahead distance $D_{LookAhead}$ along the optimal trajectory generated by the convoy controller.

The local planner takes these inputs and generates a feasible path that avoids obstacles while tracking outputs from the convoy controller. This work is based on the planner defined in (\citealp{Zhang}). The selected path is then sent to an iLQR trajectory following controller to generate the control sequence.

\begin{figure*}[t!]
     \centering
     \begin{subfigure}[b]{0.32\textwidth}
         \centering
         \includegraphics[width=\textwidth]{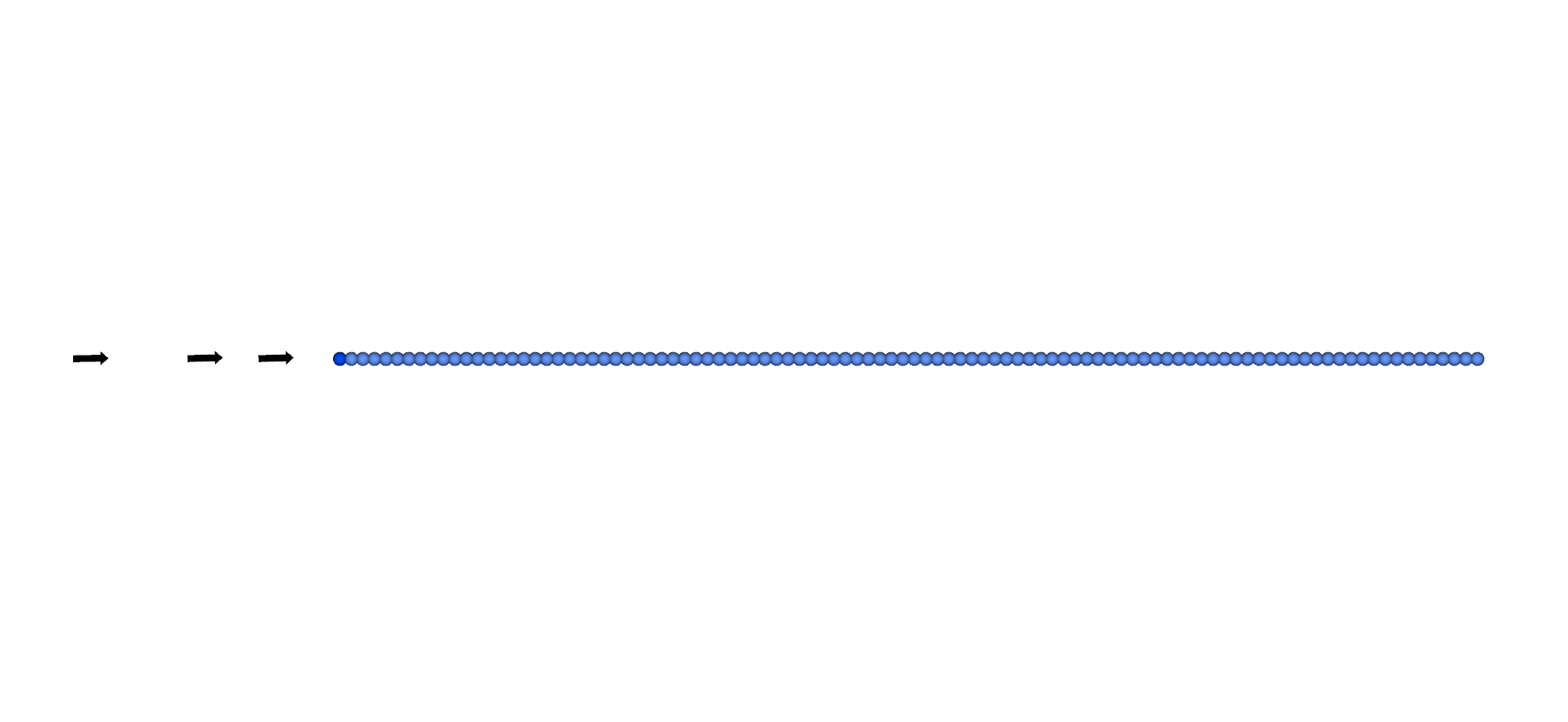}
         \caption{Straight Line}
         \label{fig:straight_line}
     \end{subfigure}
     \hfill
     \begin{subfigure}[b]{0.32\textwidth}
         \centering
         \includegraphics[width=\textwidth]{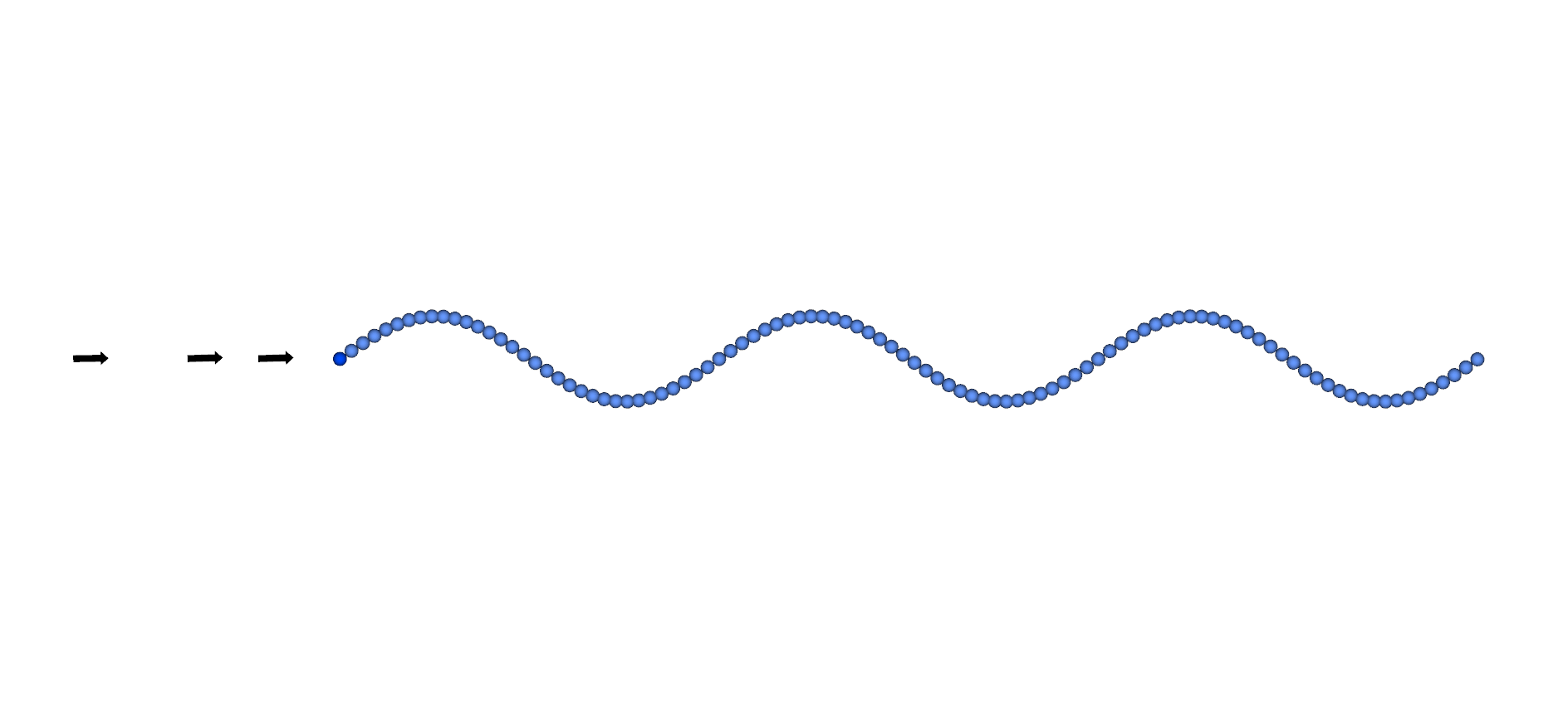}
         \caption{Low Curvature}
         \label{fig:low_curvature}
     \end{subfigure}
     \hfill
     \begin{subfigure}[b]{0.32\textwidth}
         \centering
         \includegraphics[width=\textwidth]{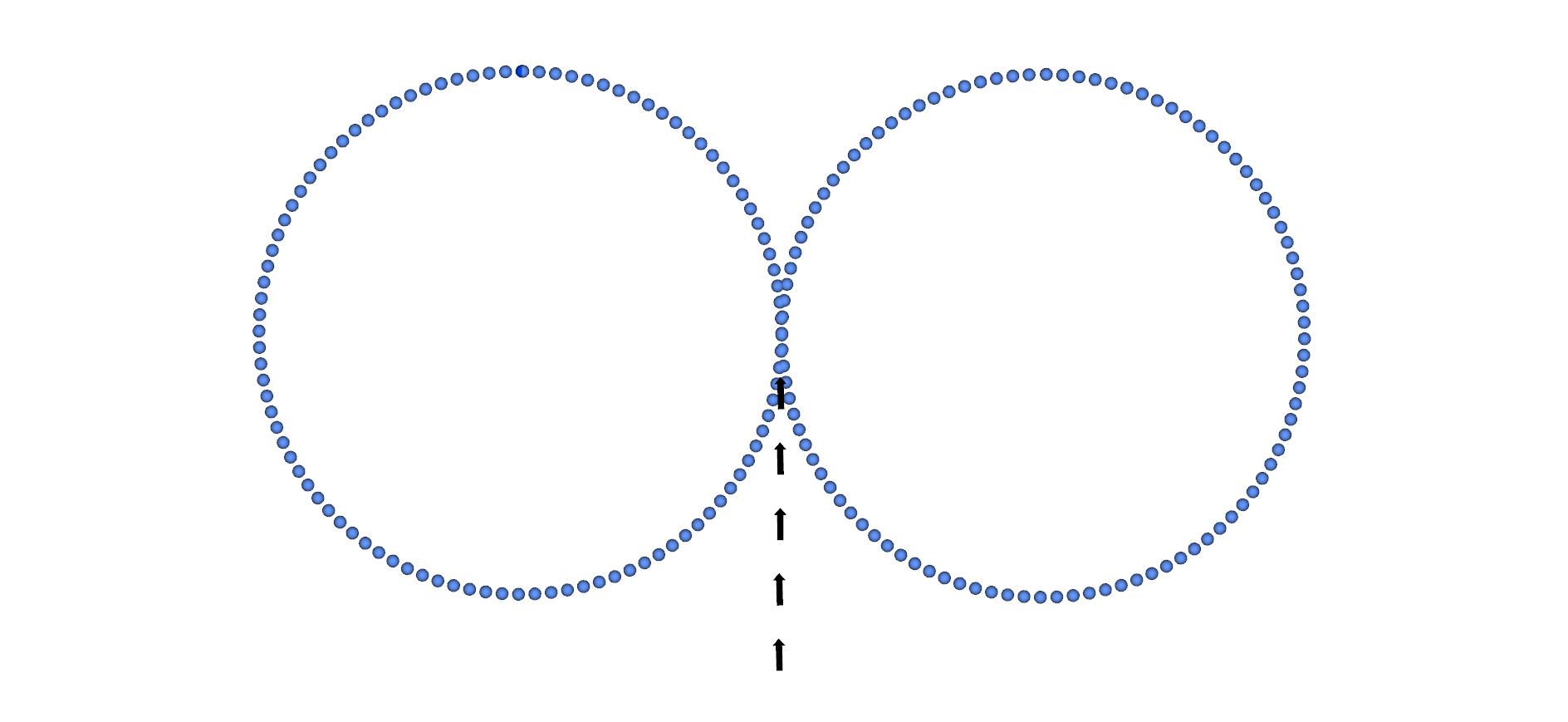}
         \caption{High speed $\infty$ loop}
         \label{fig:SimFig8}
     \end{subfigure}
     \hfill
     \begin{subfigure}[b]{0.32\textwidth}
         \centering
         \includegraphics[width=\textwidth]{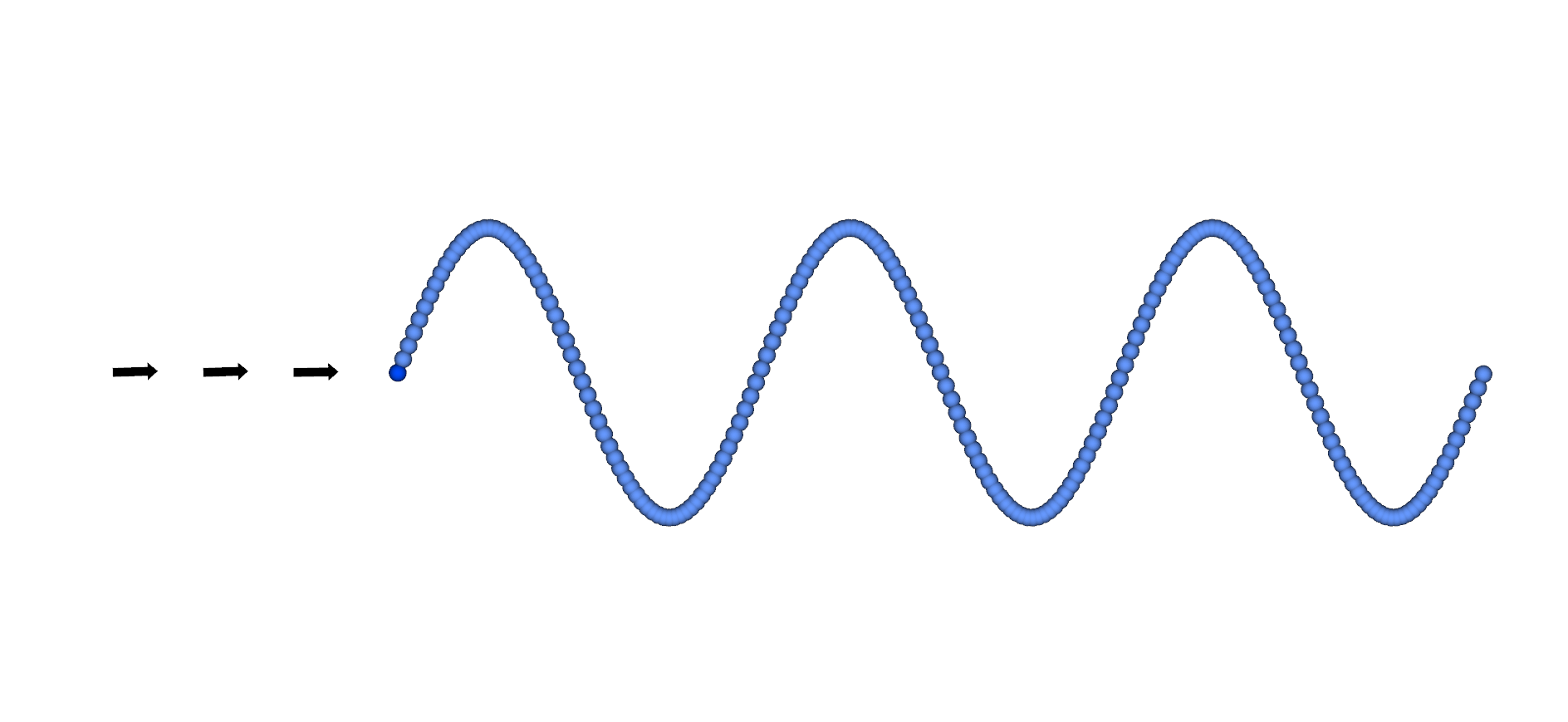}
         \caption{Tight turns}
         \label{fig:Sim3Sine}
     \end{subfigure}
     \hfill
     \begin{subfigure}[b]{0.32\textwidth}
         \centering
         \includegraphics[width=\textwidth]{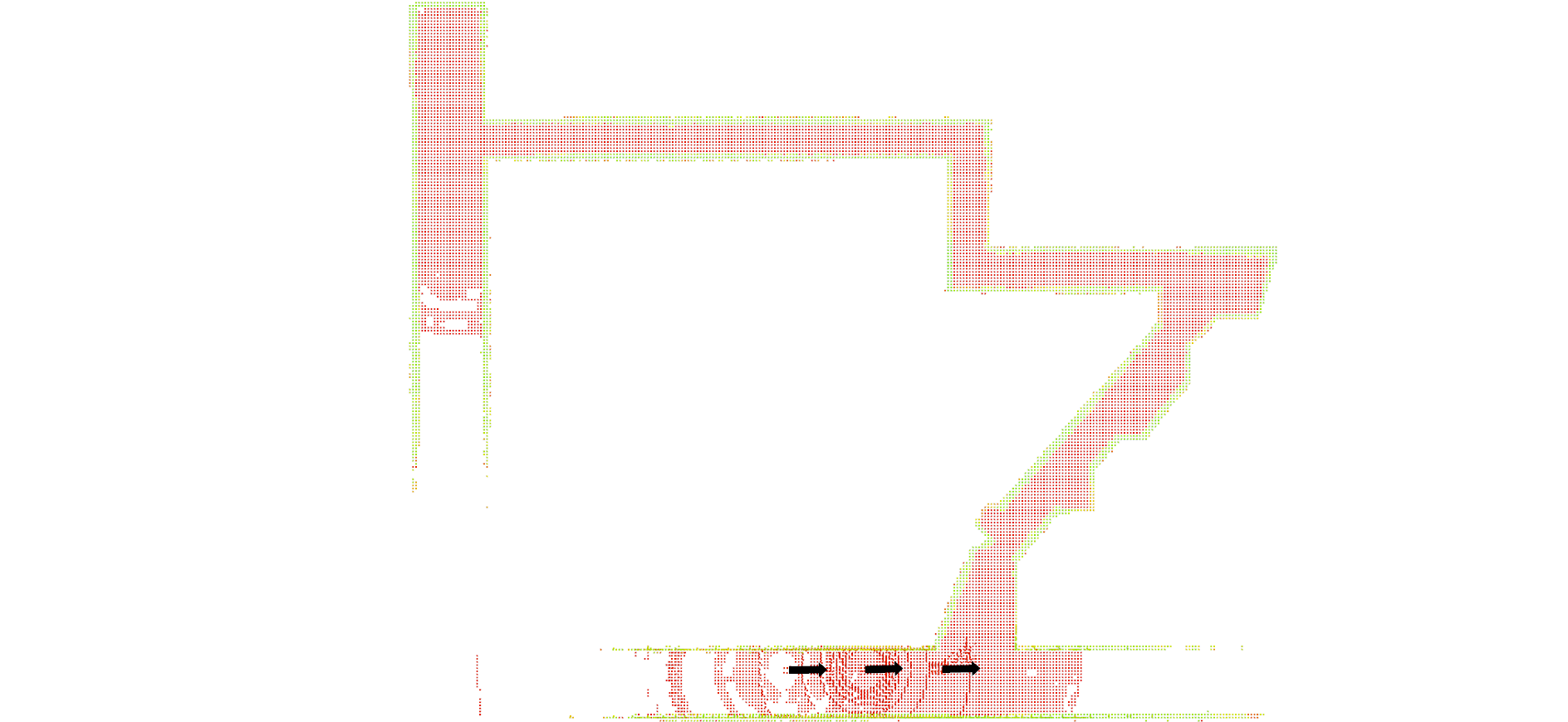}
         \caption{Tunnel environment}
         \label{fig:SimAdvTunnel}
     \end{subfigure}
     \hfill
     \begin{subfigure}[b]{0.32\textwidth}
         \centering
         \includegraphics[width=\textwidth]{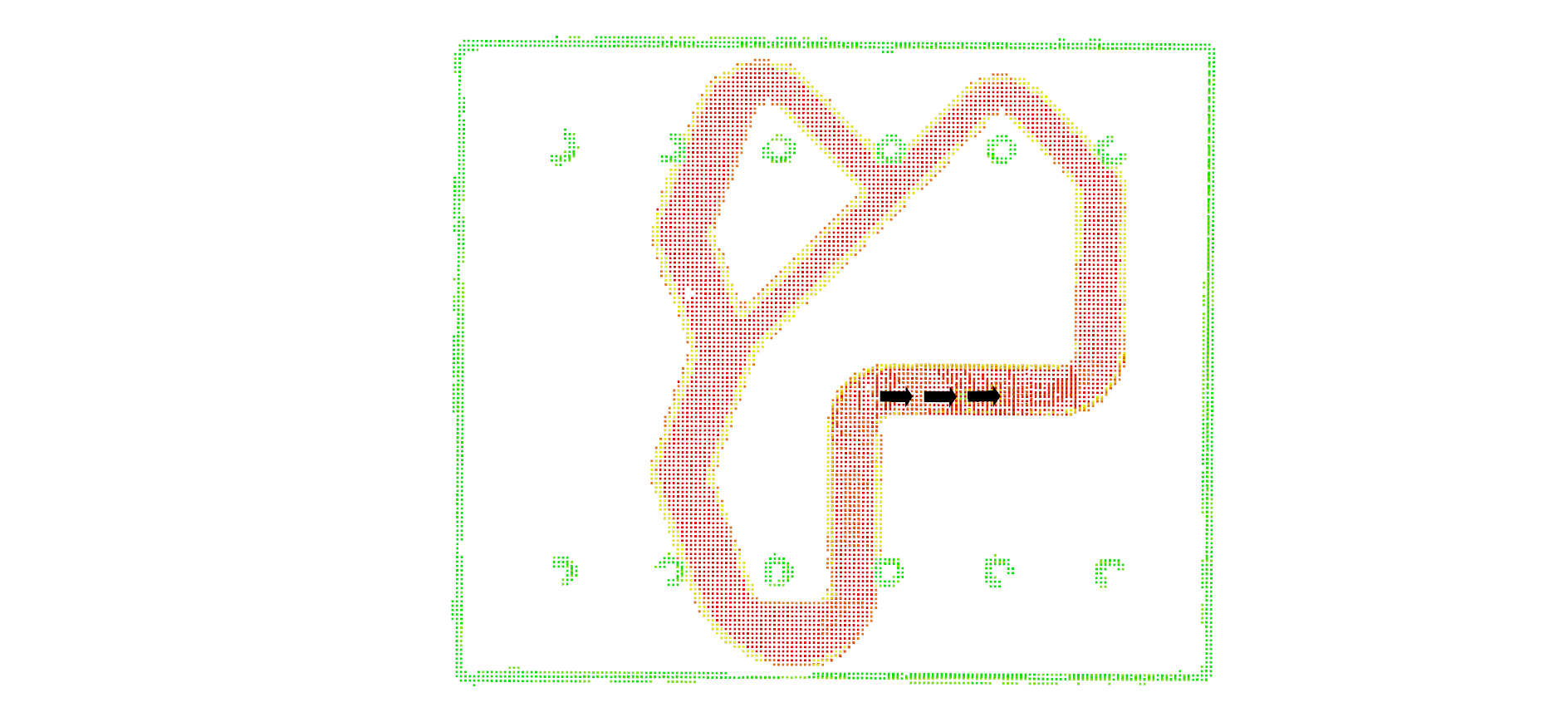}
         \caption{Race track}
         \label{fig:SimRacetrackGaz}
     \end{subfigure}
        \caption{\centering Simulation Environments}
        \label{fig:sim envs}
\end{figure*}

\begin{figure}[h]
     \centering
     \begin{subfigure}[b]{0.24\textwidth}
         \centering
         \includegraphics[width=\textwidth]{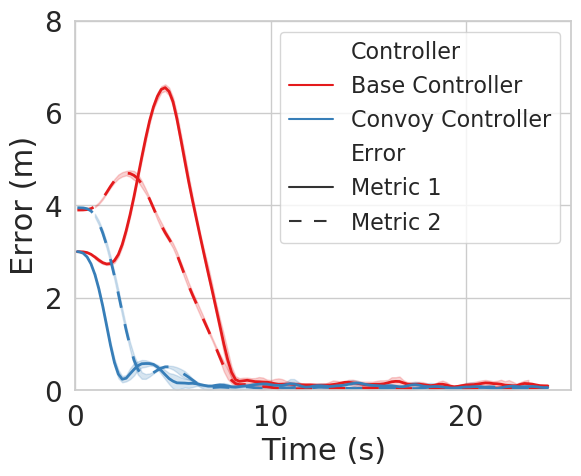}
         \caption{Straight Line}
         \label{straight_line}
     \end{subfigure}
     \begin{subfigure}[b]{0.24\textwidth}
         \centering
         \includegraphics[width=\textwidth]{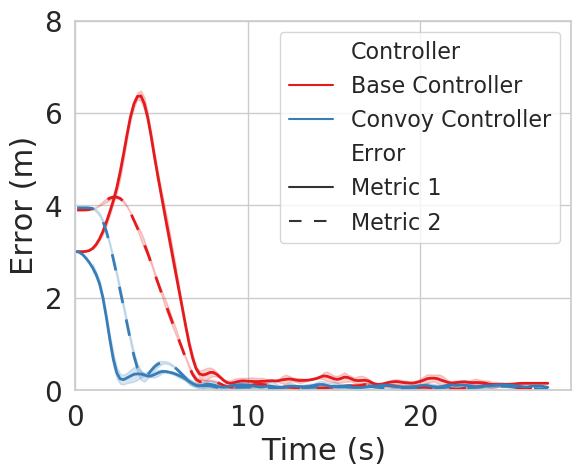}
         \caption{Low Curvature}
         \label{low_curvature}
     \end{subfigure}
        \caption{\centering Simulation Results}
        \label{fig:sim_results}
\end{figure}

\begin{figure*}[h]
     \centering
     \begin{subfigure}[b]{0.19\textwidth}
         \centering
         \includegraphics[width=\textwidth]{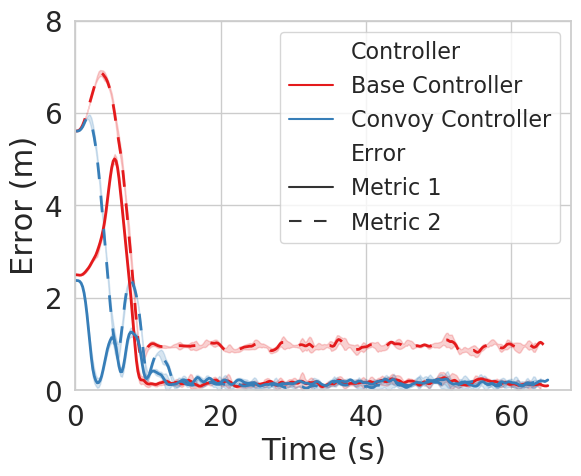}
         \caption{4 m/s}
         \label{speed4}
     \end{subfigure}
     \begin{subfigure}[b]{0.19\textwidth}
         \centering
         \includegraphics[width=\textwidth]{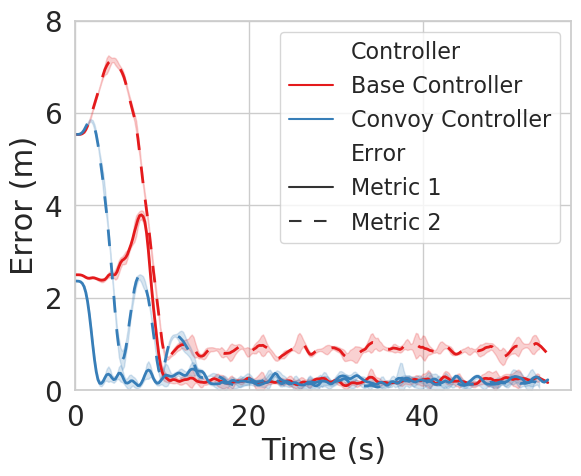}
         \caption{5 m/s}
         \label{speed5}
     \end{subfigure}
     \begin{subfigure}[b]{0.19\textwidth}
         \centering
         \includegraphics[width=\textwidth]{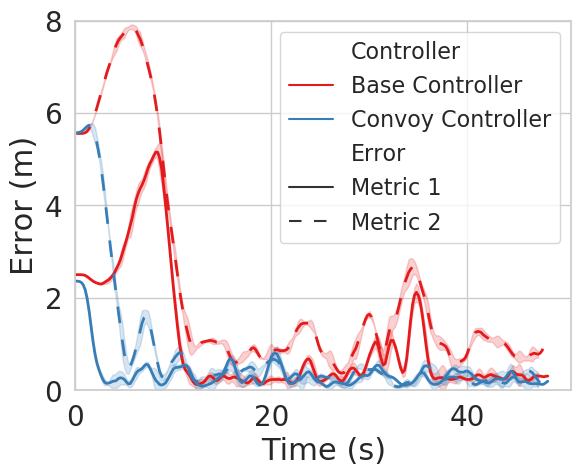}
         \caption{6 m/s}
         \label{speed6}
     \end{subfigure}
     \begin{subfigure}[b]{0.19\textwidth}
         \centering
         \includegraphics[width=\textwidth]{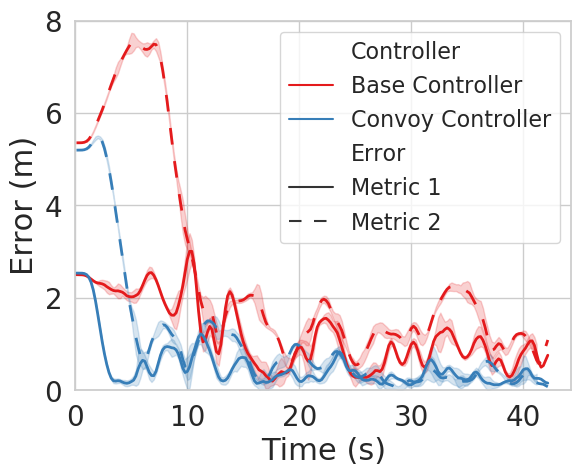}
         \caption{7 m/s}
         \label{speed7}
     \end{subfigure}
     \begin{subfigure}[b]{0.19\textwidth}
         \centering
         \includegraphics[width=\textwidth]{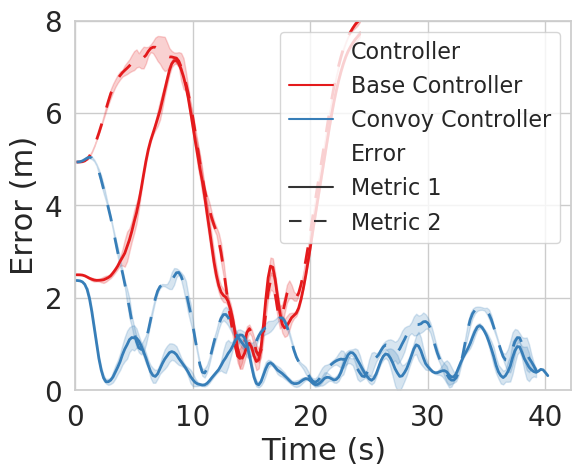}
         \caption{8 m/s}
         \label{speed8}
     \end{subfigure}        
        \caption{\centering Speed Variations in an $\infty$ loop}
        \label{fig:speedresult}
\end{figure*}

\section{Simulation Results} \label{sec:Simulation}

The performance of the presented controller (``Convoy Controller'') is characterized in a variety of simulated environments shown in Fig.~\ref{fig:sim envs}. The underlying simulator is Gazebo (\citealp{gazebo2004}). As discussed in Section \ref{LitReview}, there are many different approaches to enforcing convoy structure at the control and planning level. To develop a comparison between the presented methodology and existing literature, a baseline ``Base Controller'' is developed by combining the local planning and distance variation behaviors from (\citealp{Zhao}) with additional modifications from (\citealp{Bayuwindra,Albrecht,Ahmed}). This combination creates a decentralized controller, similar to our proposed method, and outperforms the individual performance of each work separately with respect to minimizing inter-agent distances without breaking convoy formation.

Two error metrics are used to compare the results between both controller performances. The first metric is as follows:
\begin{equation*}
    e_{m_1}= 
\begin{cases}
    abs((d_{i-1, i+1}/2) - d_{i-1, i}) & i \in (1, L)\\\
    abs((d_{L-2, L}/2) - d_{L-1, L}) & i = L
\end{cases}
\end{equation*}

This metric aims to understand how well a robot is able to maintain a position midway between its adjacent robots. Variation in this metric can help understand accordion-like behaviors that negatively affect fleet performance with continuous acceleration and braking requirements. However, this error metric can maintain a low value with all robots maintaining an equally large following distance, which is not desirable. To that end, we define a second error metric as follows:
\begin{equation*}
    e_{m_2}= 
    \text{abs}(d_{i-1, i} - d_{desired})
\end{equation*}
This metric measures distance from the desired gap between robots and, along with the first metric, can provide a thorough understanding of the convoy performance.


\begin{figure}[h]
     \centering
     \begin{subfigure}[b]{0.2\textwidth}
         \centering
         \includegraphics[width=\textwidth]{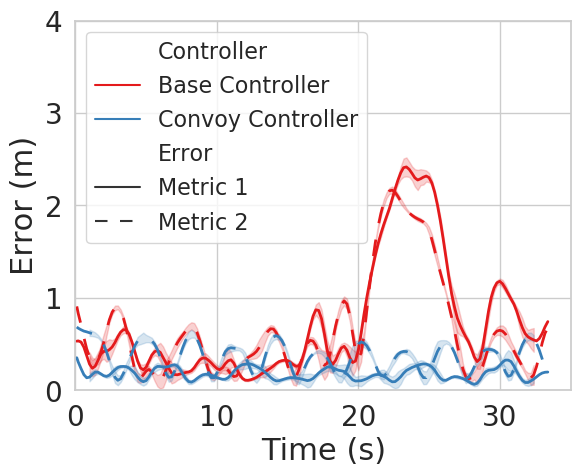}
         \caption{Tight Turns}
         \label{AvgErr3sin}
     \end{subfigure}
     \begin{subfigure}[b]{0.205\textwidth}
         \centering
         \includegraphics[width=\textwidth]{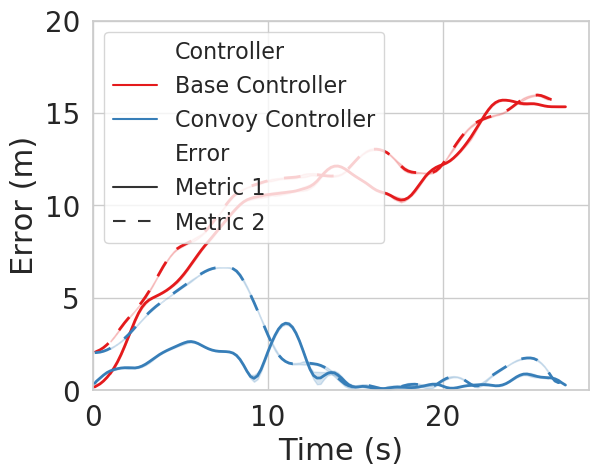}
         \caption{Tunnel}
         \label{AvgErrTunnel}
     \end{subfigure}
     \hfill
     \begin{subfigure}[b]{0.2\textwidth}
         \centering
         \includegraphics[width=\textwidth]{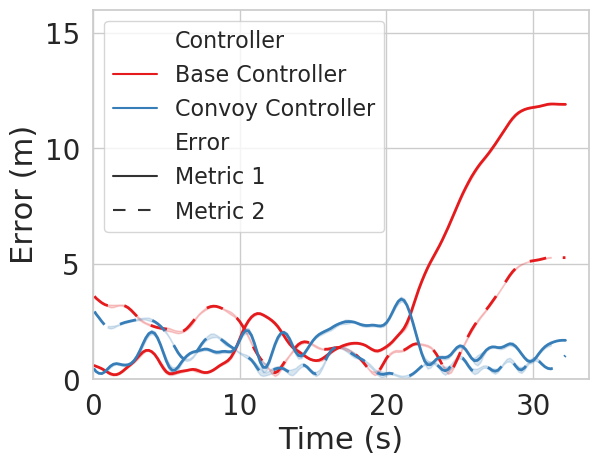}
         \caption{Race Track}
         \label{AvgErrRacetrack}
     \end{subfigure}
        \caption{\centering Simulation Results - Edge Cases}
        \label{fig:sim_envs}
\end{figure}

An initial comparison is done on a straight line and low curvature turn path. Next, an $\infty$ loop with a 20m radius is simulated at speeds varying between 4 and 8 m/s in order to test controller performance in continuous turns. Finally, runs are performed on tight turns, a tunnel environment and a racetrack. Additional constraints on agent operability and speed are added on these runs to understand fleet performance in edge scenarios.

\subsection{Straight line and low curvature runs} \label{subsec:simstraight}
The first set of simulations was performed on a group of robots operating on simple paths while starting from scattered initial locations. The aim of this experiment is to understand how quickly various controllers settle and how the error values vary during that period. As can be seen in Fig.~\ref{fig:sim_results}, our convoy controller is able to settle faster while maintaining lower values on both error metrics. This displays faster convoy structure formation through our framework, as each robot has a cost associated with all of its adjacent agents.

\subsection{High speed $\infty$ loop}
We run robots in an $\infty$ loop at various speeds to understand system capability limits. Fig.~\ref{fig:speedresult} provides error metric graphs run at various target speeds between 4 - 8 m/s. It can be seen through the figures that at initial speeds, error metrics for both controllers settle, while our control achieves this faster. However, as speeds increase, the difference in performance between our convoy controller and the base controller increases until a point when the base controller breaks and isn't able to maintain a convoy structure.

\subsection{Tight turns}
A series of waypoints are sent to the lead robot, the other robots have no information about these points. On running the base controller, the following agents tend to overshoot due to the presence of sharp turns. This behavior is seen even if we tune down the look-ahead value to zero~(\citealp{Bayuwindra}), depicting an inherent unwillingness to follow through with such sharp turns at high speeds. The error metric comparison between controllers can be seen in Fig.~\ref{AvgErr3sin}. Our controller performs better where usual convoying techniques overshoot and has difficulty tracking such tight turns. 

\subsection{Tunnel environments}
An agent might get stuck during operation, especially in indoor environments at high speeds. To simulate this, we initially stop the first robot in place for 8 seconds and then revert to normal conditions. We simulate this through narrow tunnels. In Fig.~\ref{AvgErrTunnel}, the error metrics are capped in our approach, whereas they continuously rise with the base controller as the last robot isn't able to get back into the convoy. Due to the cost term corresponding to the immediate lead and follower in the convoy controller cost definition, as soon as one of the agents isn't maintaining desired motion, the inter-robot gaps increase and the corresponding cost term shoots up. This causes the remaining agents to slow down and come to a stop until the slowed-down follower agent rejoins, capping the error metric.

\subsection{Race track}
We simulate a constrained outdoor environment at 4~m/s. We also add an additional constraint that one of the robots isn't able to achieve speeds higher than 3 m/s, which would mimic a robot with operational issues or a heterogeneous set of robots. We run this setup on a race track, shown in Fig. \ref{fig:SimRacetrackGaz}. The error metrics are seen in Fig.~\ref{AvgErrRacetrack}. The initial performance between the controllers is similar until the point where the leader makes a turn after the straight section on the race track. On the straight section, the base controller notices a continuous drop in performance that it is unable to recover past a turn, and the slow robot can no longer track the lead along the race track. Our system, on the other hand, adapts to this robot and, despite losing a little performance, is able to maintain an upper limit on the error metric and continue on the desired path.

In summary, addressing all requirements from the problem definition in Section \ref{sec:probstatement}, our convoy controller achieves lower inter-robot distances, quickly adapts to variations in environments, and achieves faster convergence for error metrics when compared with the base controller.

\begin{figure}
     \centering
     \begin{subfigure}[b]{0.35\textwidth}
         \centering
         \includegraphics[width=\textwidth]{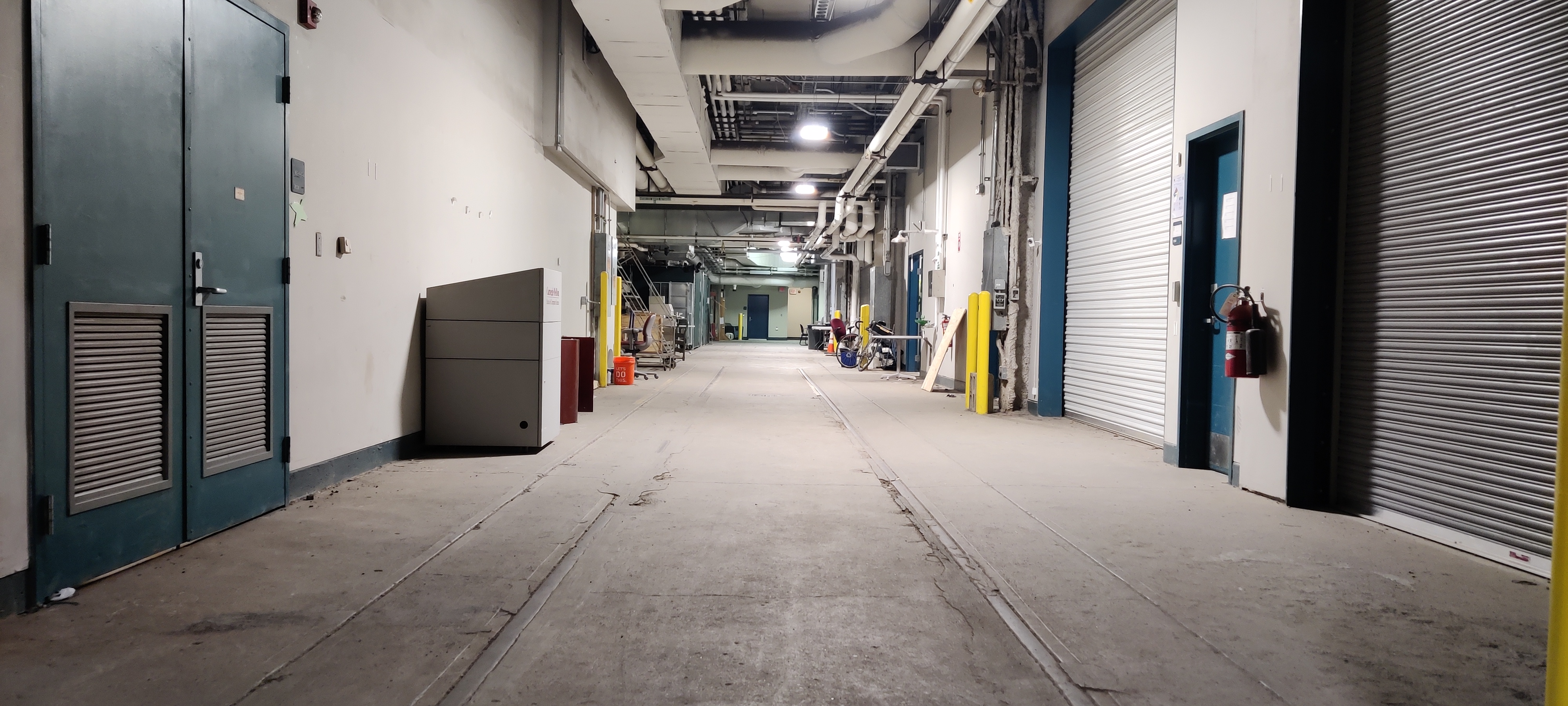}
         \caption{Straight Line}
         \label{StraightPic}
     \end{subfigure}
     \hfill
     \begin{subfigure}[b]{0.35\textwidth}
         \centering
         \includegraphics[width=\textwidth]{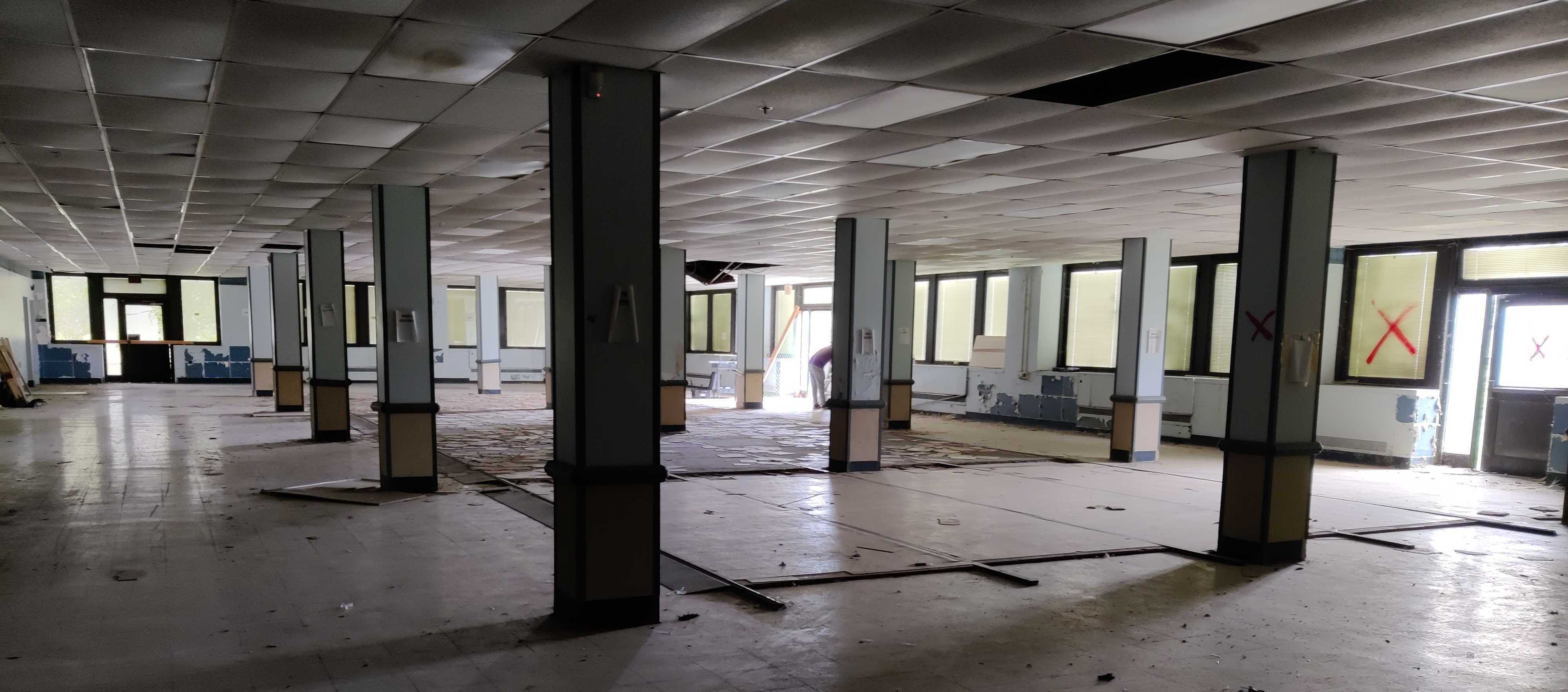}
         \caption{Column Room}
         \label{ColumnPic}
     \end{subfigure}
     \hfill
     \begin{subfigure}[b]{0.35\textwidth}
         \centering
         \includegraphics[width=\textwidth]{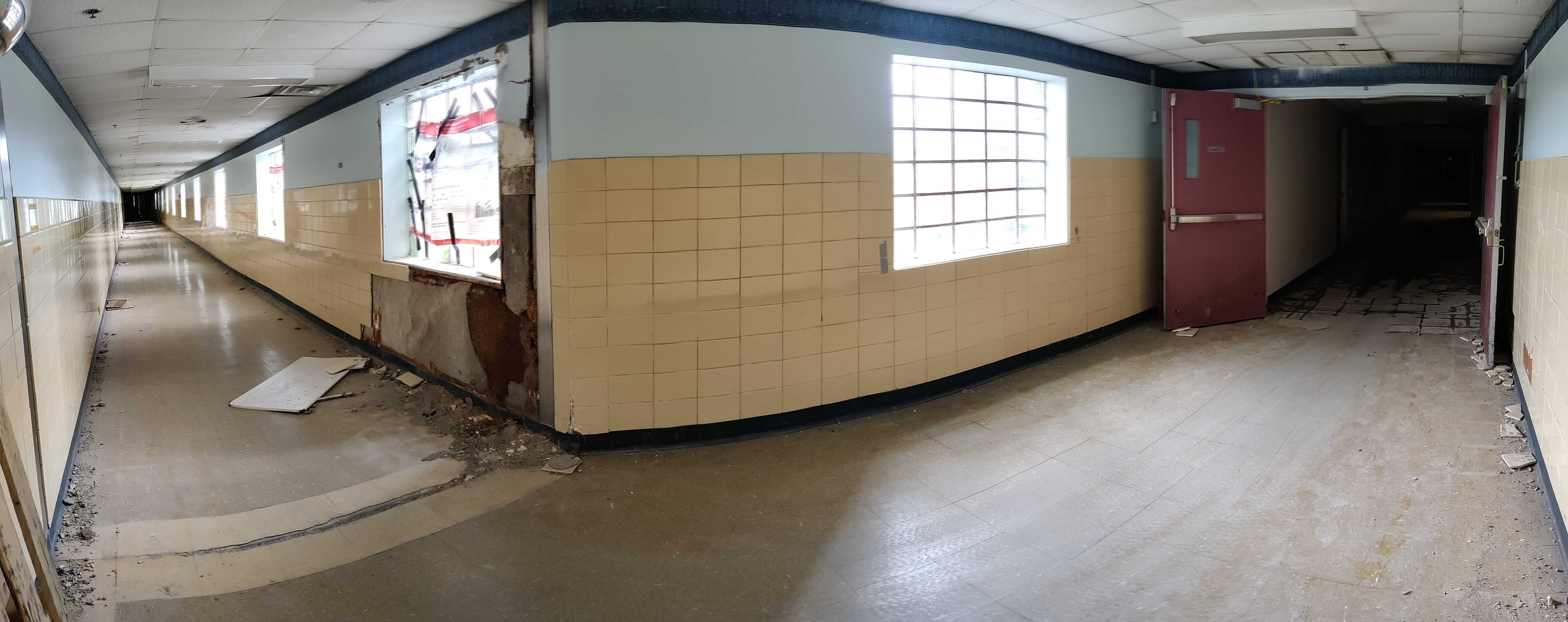}
         \caption{Long Corridor}
         \label{CorridorPic}
     \end{subfigure}
        \caption{\centering Test Environments}
        \label{fig:testenv}
\end{figure}

\begin{figure}
     \centering
     \begin{subfigure}[b]{0.12\textwidth}
         \centering
         \includegraphics[width=0.95\textwidth]{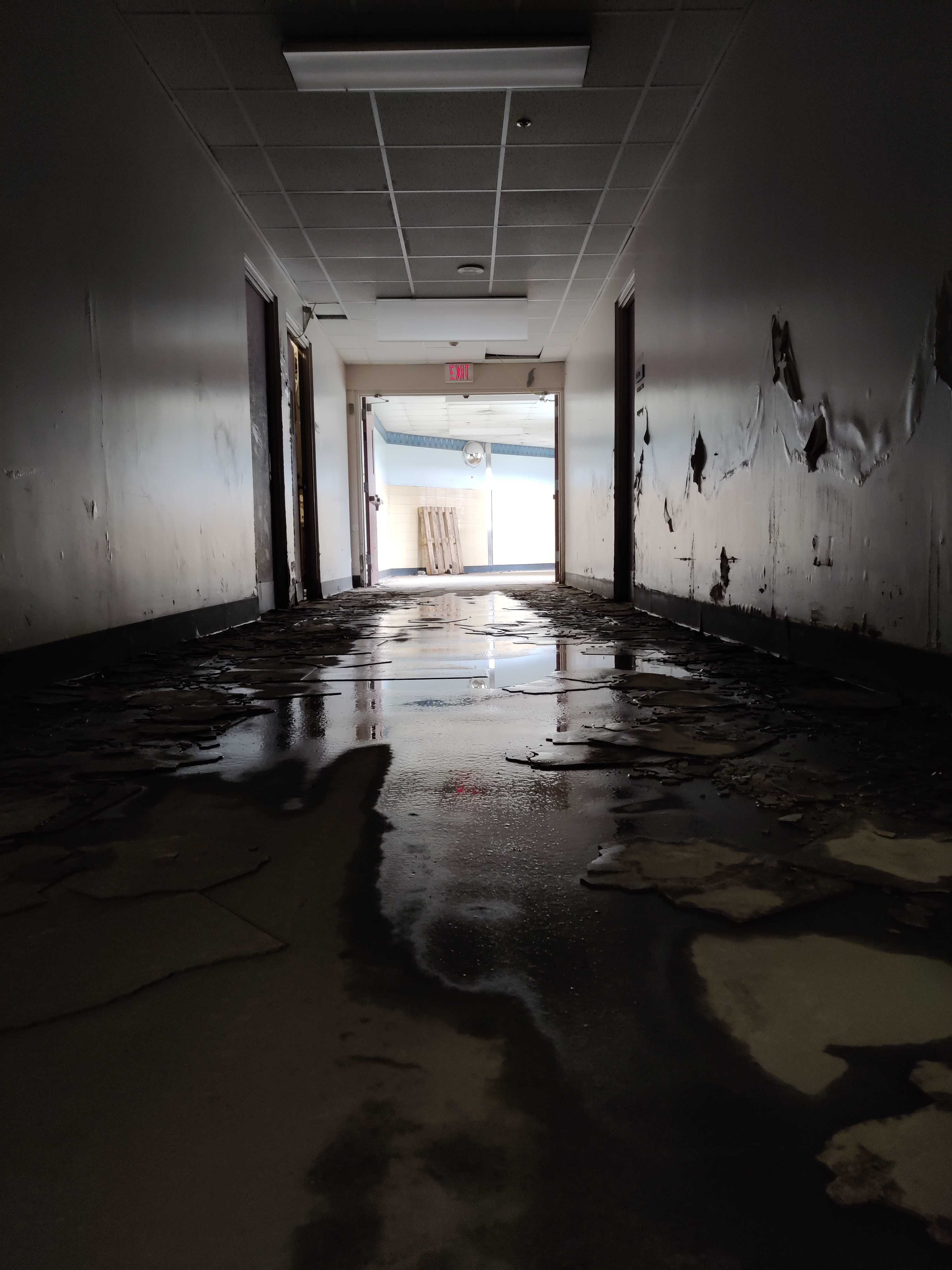}
         \caption{Puddles}
         \label{Debris1}
     \end{subfigure}
     \begin{subfigure}[b]{0.12\textwidth}
         \centering
         \includegraphics[width=0.95\textwidth]{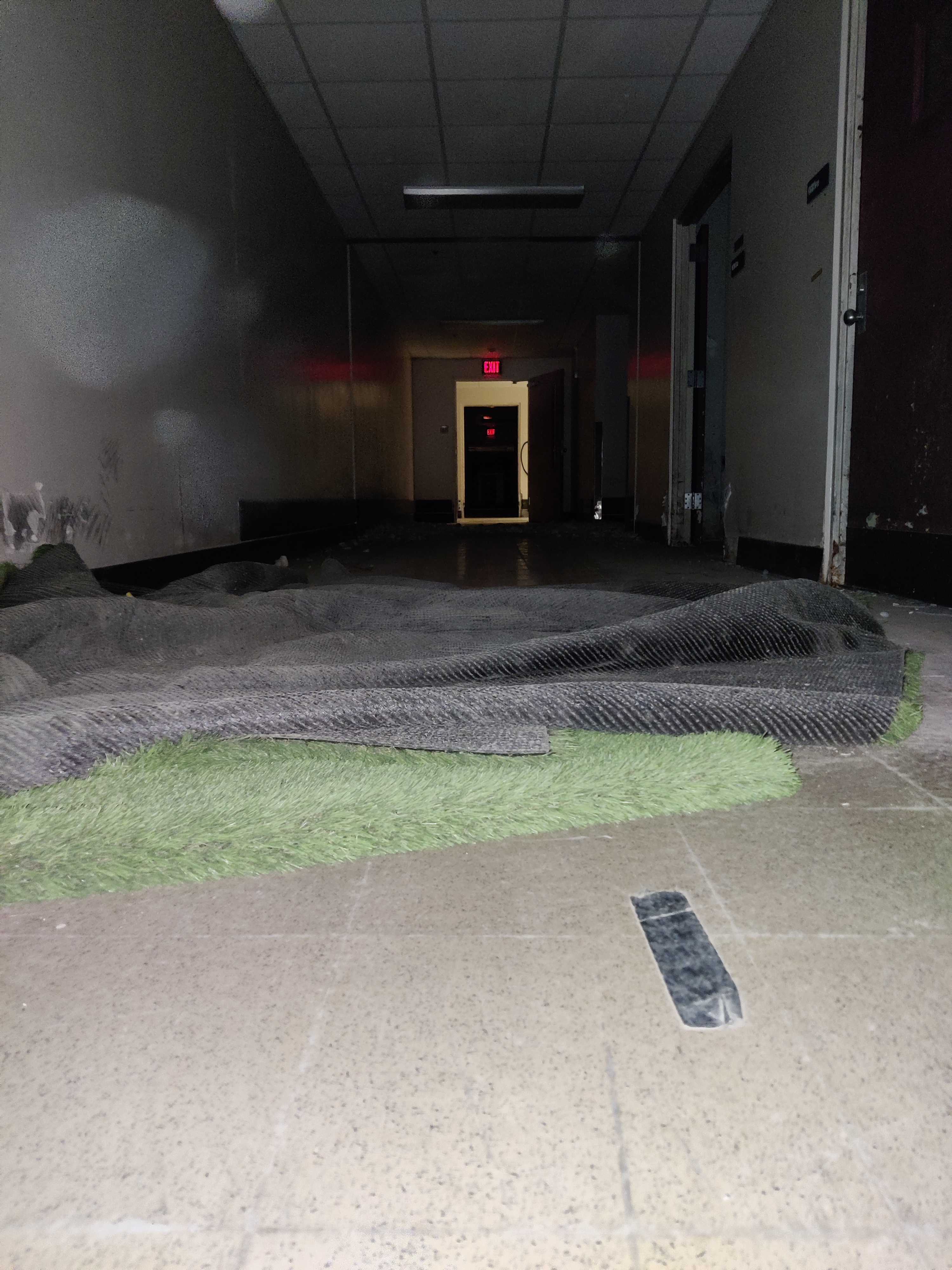}
         \caption{Carpets}
         \label{Debris2}
     \end{subfigure}
     \begin{subfigure}[b]{0.12\textwidth}
         \centering
         \includegraphics[width=0.95\textwidth]{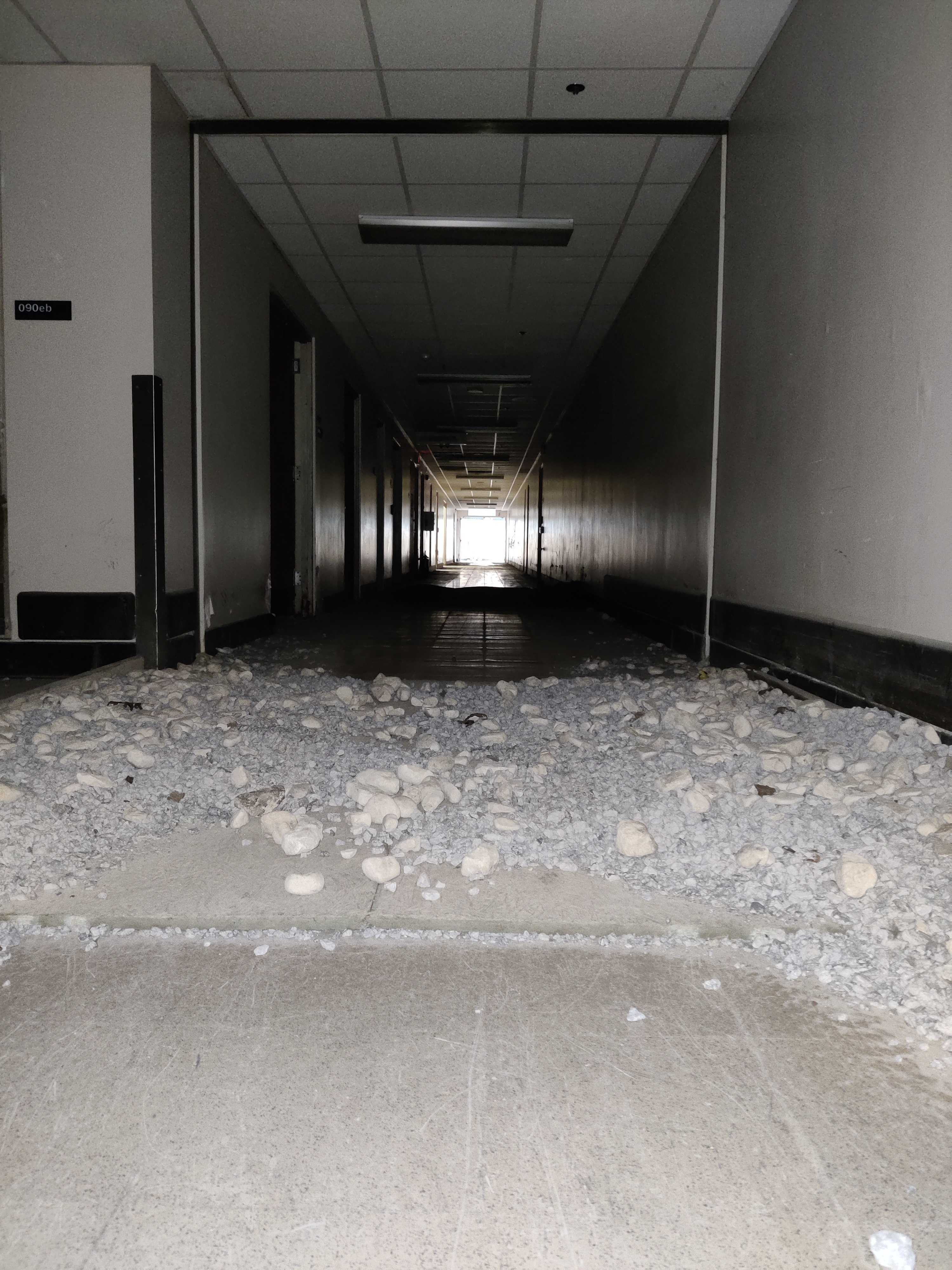}
         \caption{Stones}
         \label{Debris3}
     \end{subfigure}
     \begin{subfigure}[b]{0.12\textwidth}
         \centering
         \includegraphics[width=0.95\textwidth]{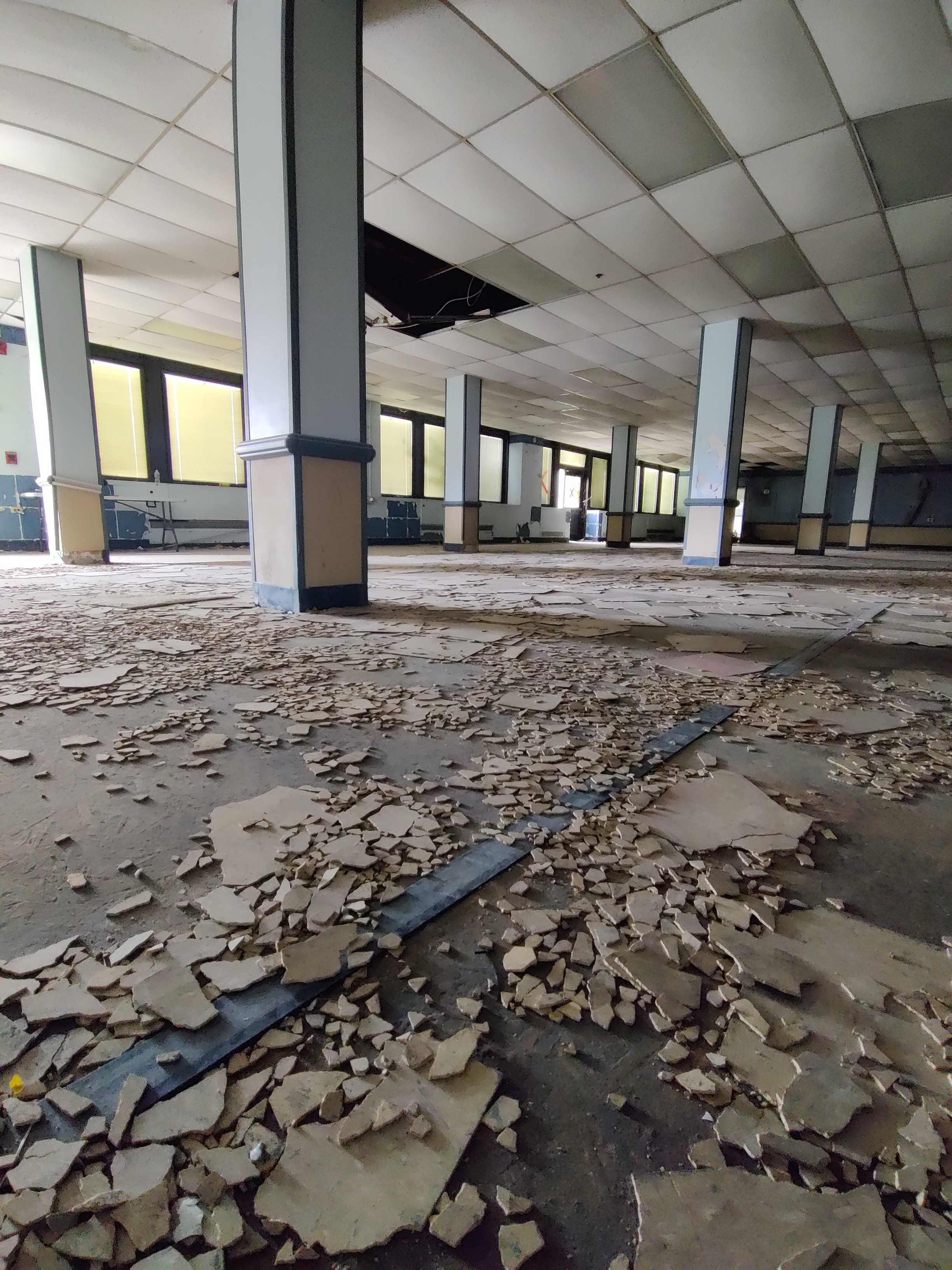}
         \caption{Broken Floor}
         \label{Debris4}
     \end{subfigure}
        \caption{\centering Ground Conditions}
        \label{fig:ground}
\end{figure}

\begin{table*}[t!]
\centering

\begin{tabular}{|c|cc||cc|}
\hline
                                       & \multicolumn{2}{c||}{\textbf{Average Error Metric 1 (m)}}                                                        & \multicolumn{2}{c|}{\textbf{Average Error Metric 2 (m)}}                                                        \\ \cline{2-5} 
{\textbf{Environment}} & \multicolumn{1}{c|}{\textbf{Base Controller}}              & \textbf{Our Controller}              & \multicolumn{1}{c|}{\textbf{Base Controller}}              & \textbf{Our Controller}              \\ \hline
\textbf{Straight Line}                 & \multicolumn{1}{c|}{{\color[HTML]{EA4335} \textbf{1.35}}}  & {\color[HTML]{34A853} \textbf{0.31}} & \multicolumn{1}{c|}{{\color[HTML]{EA4335} \textbf{1.08}}}  & {\color[HTML]{34A853} \textbf{0.44}} \\ \hline
\textbf{Sine Curve}                    & \multicolumn{1}{c|}{{\color[HTML]{EA4335} \textbf{1.12}}}  & {\color[HTML]{34A853} \textbf{0.30}} & \multicolumn{1}{c|}{{\color[HTML]{EA4335} \textbf{0.80}}}  & {\color[HTML]{34A853} \textbf{0.47}} \\ \hline
\textbf{$\infty$ loop}                   & \multicolumn{1}{c|}{{\color[HTML]{EA4335} \textbf{1.69}}}  & {\color[HTML]{34A853} \textbf{0.32}} & \multicolumn{1}{c|}{{\color[HTML]{EA4335} \textbf{2.82}}}  & {\color[HTML]{34A853} \textbf{0.73}} \\ \hline
\textbf{Tight Turn}                    & \multicolumn{1}{c|}{{\color[HTML]{EA4335} \textbf{0.72}}}  & {\color[HTML]{34A853} \textbf{0.18}} & \multicolumn{1}{c|}{{\color[HTML]{EA4335} \textbf{0.65}}}  & {\color[HTML]{34A853} \textbf{0.37}} \\ \hline
\textbf{Tunnel}                        & \multicolumn{1}{c|}{{\color[HTML]{EA4335} \textbf{10.20}}} & {\color[HTML]{34A853} \textbf{1.08}} & \multicolumn{1}{c|}{{\color[HTML]{EA4335} \textbf{10.94}}} & {\color[HTML]{34A853} \textbf{2.24}} \\ \hline
\textbf{Race Track}                    & \multicolumn{1}{c|}{{\color[HTML]{EA4335} \textbf{3.77}}}  & {\color[HTML]{34A853} \textbf{1.45}} & \multicolumn{1}{c|}{{\color[HTML]{EA4335} \textbf{2.29}}}  & {\color[HTML]{34A853} \textbf{1.02}} \\ \hline
\textbf{Column Room}                   & \multicolumn{1}{c|}{{\color[HTML]{EA4335} \textbf{1.91}}}  & {\color[HTML]{34A853} \textbf{0.97}} & \multicolumn{1}{c|}{{\color[HTML]{EA4335} \textbf{5.61}}}  & {\color[HTML]{34A853} \textbf{0.95}} \\ \hline
\textbf{Long Corridor}                 & \multicolumn{1}{c|}{{\color[HTML]{EA4335} \textbf{3.78}}}  & {\color[HTML]{34A853} \textbf{0.89}} & \multicolumn{1}{c|}{{\color[HTML]{EA4335} \textbf{5.14}}}  & {\color[HTML]{34A853} \textbf{0.68}} \\ \hline
\end{tabular}

\label{ResultsTable}
\caption{Overall results comparison}
\end{table*}

\section{Hardware Implementation} \label{sec:experiments}
\subsection{System Overview}

Each robot in our setup (Fig.~\ref{fig:conv_rubble}) consists of Traxxas remote-controlled trucks fitted with a communication node, LiDAR and IMU sensors, a Jetson AGX Xavier and a motor controller. 

For our perception framework, we make use of Super Odometry \citealp{SuperOdom}, which is an IMU-centric pipeline that provides estimates of each agent's odometry. To ensure neighboring robots don't create drifts in odometry in low LiDAR feature regions such as narrow corridors, the LiDAR data was filtered using LiDAR masks in the directions of the forward and rear robots. This ensures robust perception performance in all environments.

A base station is used to provide input commands to the convoy leader, either via a joystick or through a waypoint-sharing interface. All systems are run over ROS and use the DDS protocol for real-time inter-robot communication.

\begin{figure}
     \centering
     \begin{subfigure}[b]{0.24\textwidth}
         \centering
         \includegraphics[width=\textwidth]{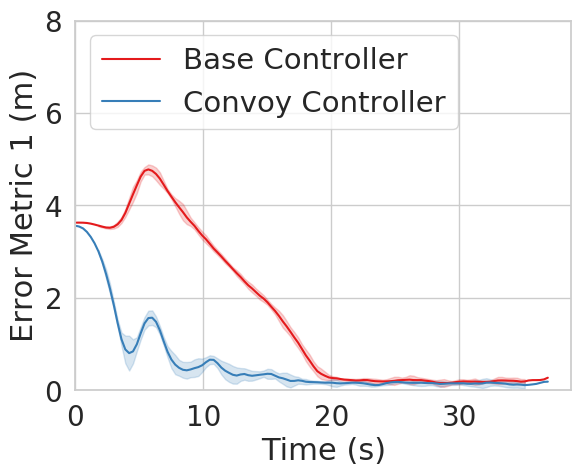}
         \caption{Metric 1 - Straight Line}
         \label{test_straight1}
     \end{subfigure}
     \hfill
     \begin{subfigure}[b]{0.24\textwidth}
         \centering
         \includegraphics[width=\textwidth]{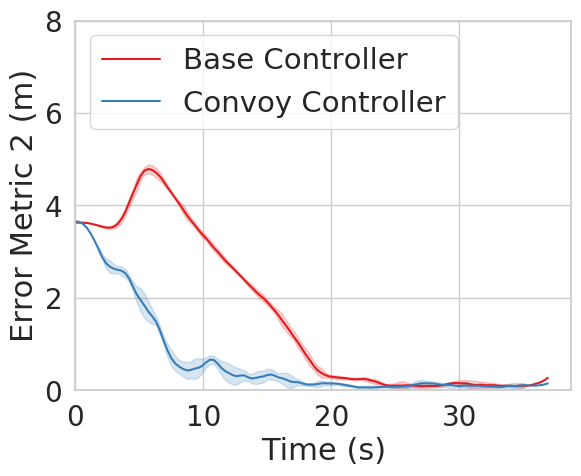}
         \caption{Metric 2 - Straight Line}
         \label{test_straight2}
     \end{subfigure}
     \hfill
     \begin{subfigure}[b]{0.24\textwidth}
         \centering
         \includegraphics[width=\textwidth]{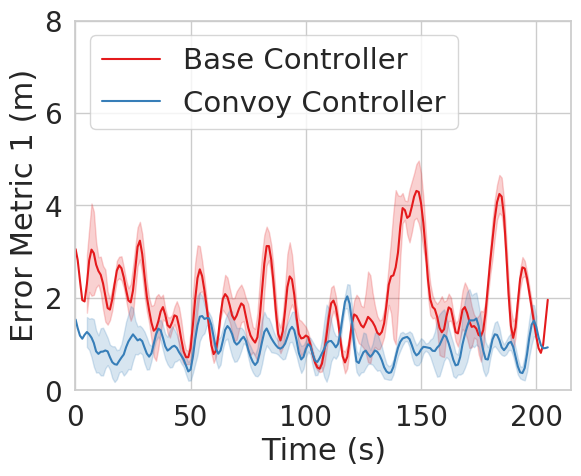}
         \caption{Metric 1 - Column Room}
         \label{AvgErrColumn}
     \end{subfigure}
     \hfill
     \begin{subfigure}[b]{0.24\textwidth}
         \centering
         \includegraphics[width=\textwidth]{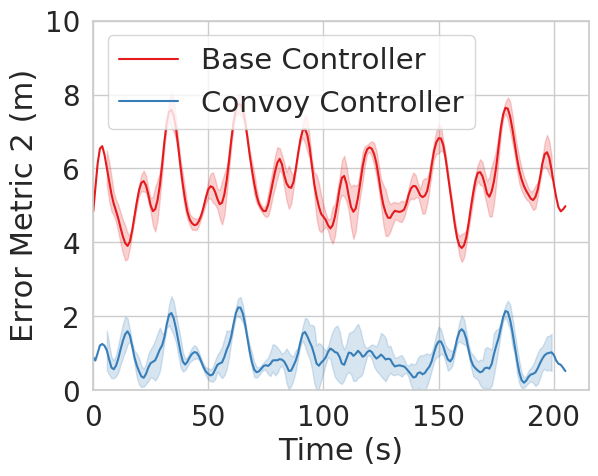}
         \caption{Metric 2 - Column Room}
         \label{AvgGapColumn}
     \end{subfigure}
     \begin{subfigure}[b]{0.24\textwidth}
         \centering
         \includegraphics[width=\textwidth]{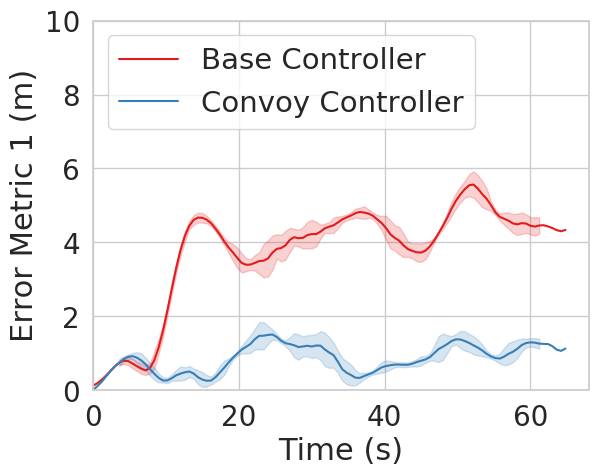}
         \caption{Metric 1 - Long Corridor}
         \label{AvgErrCorridor}
     \end{subfigure}
     \hfill
     \begin{subfigure}[b]{0.24\textwidth}
         \centering
         \includegraphics[width=\textwidth]{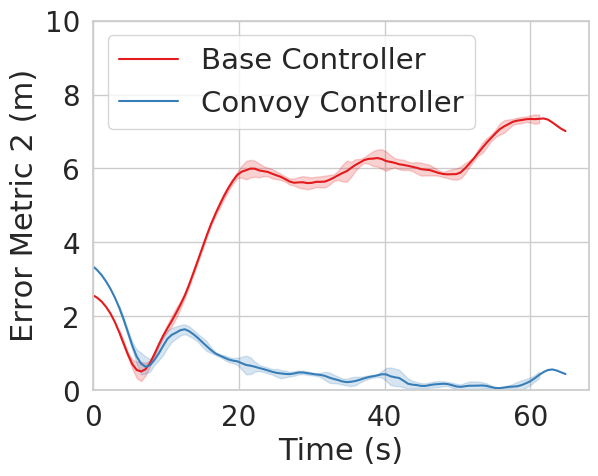}
         \caption{Metric 2 - Long Corridor}
         \label{AvgGapCorridor}
     \end{subfigure}
        \caption{\centering Test Results}
        \label{fig:test_results}
\end{figure}

\subsection{Results}

\subsubsection{Straight Line:}
As discussed in Section~\ref{subsec:simstraight}, we test similar conditions on the three ground robots. It can be seen that the error metrics in Fig.~\ref{test_straight1} and~\ref{test_straight2} that the metrics vary similarly to those in Fig.~\ref{straight_line} and~\ref{low_curvature} where our convoy controller is able to achieve steady state values faster than the existing base controller. This also re-validates simulation results on a real system.

\subsubsection{Column room:}
Such an environment (Fig.~\ref{ColumnPic}) is able to test convoy agility due to the continuous obstacle avoidance requirements via low-curvature maneuvers while maintaining a stable convoy structure during high-speed maneuvers. The room has dimensions of 35~m~x~20~m. The operator commands a general direction of motion to the lead vehicle and all three robots make use of the convoy controller to follow closely while achieving speeds of 4~m/s. The robots make continuous turns to not hit walls, columns and other obstacles. The robots run in various directions around this room to cover a total distance of over 600~m. The error metrics between our and the base controller are shown in Fig.~\ref{AvgErrColumn} and~\ref{AvgGapColumn}. Despite having to continuously accelerate, decelerate and turn in such an environment, the convoy controller is able to maintain average gaps of just over 5~m while operating at 4 m/s compared to 9~m following distances through the base controller. We also run the same path with a single robot and notice an increase in average speed by under 5\%, displaying the fact that our framework doesn't reduce overall robot capabilities while increasing agents.

\subsubsection{Long corridors:}
In narrow long corridors with doorways (Fig.~\ref{CorridorPic}), we illustrate high-speed convoy performance while adapting to turns and doorways. These environments contain debris as shown in Fig.~\ref{fig:ground}, testing robustness in our control system and mimicking search and rescue operations. The performance difference can be seen in Fig.~\ref{AvgErrCorridor} and~\ref{AvgGapCorridor}. The average inter-robot distances are just under 4~m at high speeds, which is significantly lower than the lowest gaps of 9-15~m seen at comparable speeds in other research papers (\citealp{Zhao, Shin}) and the base controller. The deviations in the metrics only take place when the robots make a turn into another corridor, move over rough grounds, or navigate through doorways. The performance drop of a convoy against a single robot is less than 2\%, indicating almost similar time to achieve the goal. The robots were able to navigate over 3~km in such environments without running into each other or leaving a robot behind, displaying a robust and safe system.

\section{Conclusions}
\label{sec:conclusions}
We have designed an optimal decentralized control system to run on a multi-robot convoy. Our design incorporates future predictions on the adjacent robots and solves a cost minimization incorporating robot states and controls, allowing the agents to operate closer to each other at high speeds. This, in turn, enables robots to operate well in environments that require continuous sharp turns and variations in speeds. The framework operates in a decentralized manner, resulting in no additional computational requirements while increasing the number of robots being deployed. We have been able to show, through simulation and hardware experimentation, the improvement in performance against the current state-of-the-art methods and have been able to cut down following distances against state-of-the-art by half. The system can potentially be improved in the future by incorporating environment-dependent control optimizations and integrating hybrid communication control methods with increased data transfer to reduce state prediction computational loads.


\bibliography{ifacconf}

\appendix
\section{Quadratic Form Expansion} \label{app1}

Given vectors $x\in\mathbb{R}^n$, $x_a\in\mathbb{R}^n$, $x_b\in\mathbb{R}^n$, $x_c\in\mathbb{R}^n$ and symmetric invertible matrices: $Q_a \in \mathbb{R}^{n \times n}$, $Q_b \in \mathbb{R}^{n \times n}$ and $Q_c \in \mathbb{R}^{n \times n}$, in the expression
\begin{equation} \label{app1_eqn: three_term_quadratic_form}
\begin{aligned}
    (x-x_a)^TQ_a(x-x_a) &+ (x-x_b)^TQ_b(x-x_b)\\  &+ (x-x_c)^TQ_c(x-x_c),
\end{aligned}
\end{equation}
the terms can be combined into a single quadratic form via expansion and the use of multivariate completing the square. Consider the expansion of a general single-term quadratic form such as those above. WLOG, this is performed for form $a$ and is given as

\begin{equation} \label{app1_eqn: quadratic_expansion}
    (x-x_a)^TQ_a(x-x_a) = x^TQ_ax - 2x_a^TQ_ax + x_a^TQ_ax_a.
\end{equation}
By performing the expansion expressed in Equation \ref{app1_eqn: quadratic_expansion} on each single-term quadratic form in \ref{app1_eqn: three_term_quadratic_form} and combining like terms, a single general quadratic form in $x$ can be generated as

\begin{equation}\label{app1_eqn: full_expression}
\begin{aligned}
    x^T(Q_a + Q_b &+ Q_c)x - 2(x_a^TQ_a + x_b^TQ_b + x_c^TQ_c)x\\ &+ (x_a^TQ_ax_a + x_b^TQ_bx_b +x_c^TQ_cx_c).
\end{aligned}
\end{equation}
Making the following simplifications:
\begin{align*}
    Q_T &= Q_a + Q_b + Q_c \\
    y_T &= Q_ax_a + Q_bx_b + Q_cx_c \\
    Z_T &= x_a^TQ_ax_a + x_b^TQ_bx_b + x_c^TQ_cx_c, 
\end{align*}
\eqref{app1_eqn: full_expression} may be equivalently expressed as
\begin{equation*}
    x^TQ_Tx - 2y_T^Tx + Z_T.
\end{equation*}
By performing a multivariate complete the square, this expression may be rearranged into a quadratic form with the term $x$ and a remainder term:
\begin{align*}
    (x-Q_T^{-1}y_T)Q_T(x-Q_T^{-1}y_T) - y_T^TQ_Ty_T + Z_T.
\end{align*}

\clearpage

\end{document}